\documentclass[acmlarge,nonacm,screen]{acmart}
\makeatletter\let\hyxmp@parse@acmart\relax\makeatother

\graphicspath{{figures/}}

\usepackage{amsmath}
\usepackage{booktabs}
\usepackage{tikz}
\usetikzlibrary{positioning,arrows.meta,fit,calc,backgrounds}
\usepackage{fontawesome5}
\usepackage{placeins}
\usepackage{animate}

\newcommand{\head}[1]{\textbf{#1}\hspace{0.5em}}

\title[REFIT]{REFIT: Recognize, Fix, and Test Wearable Sensor Placement Shifts without Labels}

\author{Bangxun Tang}
\email{bangxunt@uci.edu}
\affiliation{%
  \institution{University of California, Irvine}
  \city{Irvine}
  \state{California}
  \country{USA}
}
\renewcommand{\shortauthors}{Tang}

\begin{abstract}
We present REFIT, an input calibration for frozen activity-recognition models whose inertial sensors are worn differently at
deployment than in training. When users move a watch to the other wrist or put a strap sensor back on turned, the model
sees the same motion on changed axes. REFIT undoes such shifts without labels or
retraining. It describes them by families of axis transforms, such as reflections and rotations, and fits each family to
the user's data so that simple statistics match those of the training data. The family that removes most of the mismatch
names the shift. REFIT fixes the shift by applying the best member of that family before the frozen
model and re-estimating its normalization statistics. It tests the fixed model with a label-free accuracy
estimate and asks the user to re-wear the sensor when it is low. Experiments on real left/right sensor pairs
and on real and simulated re-attachment show that REFIT outperforms label-free test-time adaptation methods on every
dataset and restores most of the accuracy lost to re-attachment. It names injected shifts far more reliably than a confidence-based
selector. After a correction over all signed permutations of the axes, the estimate separates successful from
failed corrections.
\end{abstract}



\begin{document}
\maketitle

\section{Introduction}
\label{sec:intro}

Inertial measurement units (IMUs) in watches, phones and body-worn sensors let deep models recognize daily activities
from motion~\cite{wang2019harsurvey,chen2021survey,ordonez2016deepconvlstm}. Representations learned from unlabeled
IMU data further reduce the labeled data that such models need~\cite{xu2021limubert}. These models learn from recordings
in which every sensor sits in a fixed place and orientation. In daily use, the way a sensor is worn changes. People wear wrist sensors on either wrist and sometimes upside down~\cite{straczkiewicz2019placement}, and whether the device sits on the dominant
or the non-dominant wrist changes what its accelerometer records~\cite{buchan2020comparison}. Strap sensors are taken off and put back in another
orientation~\cite{realdisp}, and a phone is carried the other way round in a pocket. A model that learned how gravity and
rotation appear on each sensor axis then receives the same motion on different axes, and its accuracy drops
sharply~\cite{kunze2014placement,banos2014displacement,gjoreski2016wrist}.

We call this change a placement shift. Often studied as sensor displacement or sensor placement
variation~\cite{kunze2014placement,banos2014displacement}, it differs from the heterogeneity of devices and users
that mobile sensing has studied extensively~\cite{stisen2015smart,mathur2018using,xu2023practically}. It happens to the same user and the same device,
each time the device is put on again. It is discrete: a watch sits on one wrist or the other, and a box-shaped sensor fits
its strap in a small number of orientations. It is also silent, because the model keeps producing a prediction for every
window. A deployed system therefore needs to notice and undo the shift on the user's device, without labels and without
retraining, and to tell the user when it cannot.

Correcting this at deployment is difficult for three reasons. First, users do not label their data, and a model running
on a watch or phone should not be retrained for every user. Label-free test-time adaptation re-estimates
normalization statistics~\cite{adabn}, tunes normalization parameters by entropy minimization~\cite{tent}, adapts the
feature extractor~\cite{shot}, replaces the classifier by prototypes of the unlabeled data~\cite{t3a,oftta}, or refines
the output probabilities~\cite{lame}. Several of these ideas have been carried over to wearable
sensing~\cite{oftta,gong2022note,kang2023sfadapter}. None of these updates is designed to move gravity back to the axis on which
the model expects it. Second, the shift depends on the device. After a side switch, when a watch moves to the other wrist~\cite{gjoreski2016wrist}, the
accelerometer and the gyroscope change sign on different axes. Which axes change depends on the axis convention of the
device, that is, on how its axes are laid out in its case. A re-attached sensor
can land in any orientation in which its box fits the strap. Third, the deployed system does not know whether a sensor
was moved at all, what kind of shift occurred, or whether any transform of the axes can explain it. A correction forced
onto a user whose sensor did not move, or onto a shift that no axis transform explains, gains nothing and can even lower
the accuracy.

In this work, we present REFIT, a label-free input calibration that recognizes the kind of shift (identification), fixes
it in front of a frozen model (correction), and tests the fixed model before its predictions are used (verification). Our key idea is that the common ways of wearing
a sensor differently act on the raw signal as transforms of the sensor axes. Each family of such transforms can
be fitted to unlabeled data directly, without any gradient step. We call one IMU with an accelerometer and a gyroscope a
sensor unit. For each unit, REFIT compares simple statistics of the user's windows with those of the training data, such
as the mean acceleration, which gravity dominates, and the covariances of the accelerometer and the gyroscope. It then
finds the member of each family that aligns these statistics best. To recognize
the shift, REFIT fits each family on one half of the user's windows and measures how much of the mismatch it removes on
the other half. The family
that removes most names the kind of shift, and a user whose statistics already lie within the range of the training users
is left unchanged. To fix the shift, REFIT applies the best member of the named family to the windows before they reach
the model and re-estimates the normalization statistics of the model on the corrected windows. To test the fix, it estimates the
accuracy of the corrected model with average thresholded confidence (ATC), a published label-free
estimate~\cite{garg2022atc}. A low estimate means that the shift has most likely not been undone, for example because the
sensor was moved to another body location. REFIT then abstains~\cite{geifman2017selective} and asks the user to put the sensor back in place. The
whole procedure uses no labels of the user and no gradient step, and it never changes the weights of the model.

Two families cover the most frequent shifts. A side switch mirrors the sensor through the body's plane of symmetry. The
accelerometer measures an ordinary vector and flips one axis, while the gyroscope measures an axial vector, the rate of
rotation, and flips the other two. Dietary monitoring already uses this mirror to map left-hand recordings onto the right
hand~\cite{kyritsis2021eating,rouast2020oreba,heydarian2020intake}, and sports sensing uses it to align left-handed and
right-handed players~\cite{park2024silentimpact}. Recent work applies it to a pretrained eating detector at test
time~\cite{wang2026eating}. We treat this mirror as one member of the reflection family, so that the same fit can also find
the mirror on devices with other axis conventions. A re-attachment turns a unit. REFIT searches the
orientations in which a box-shaped unit can be strapped back. The re-estimated normalization statistics absorb part
of the small strap misalignment that remains because a strap never sits exactly as before. Four further families cover permutations, sign
flips, gains and offsets of the channels. They describe changes of axis order, polarity, sensitivity and zero level,
for example after a device is replaced, and let identification tell a placement shift from such changes.

We evaluate REFIT on ten public datasets with real left/right sensor pairs, with real re-attachment, and with
simulated re-attachment of held-out subjects, against label-free adaptation methods. Under each of these shifts, it
outperforms every adaptation method on every dataset. It restores most of the accuracy lost to re-attachment and costs about as
much computation as re-estimating normalization statistics (Appendix~\ref{app:cost}). On shifts
injected into real recordings, it names the kind of shift far more often than a selector guided by the confidence of
the model. Applying the correction of the named family is more accurate over all test cases than always correcting a
rotation whatever the shift, than this confidence selector and than every adaptation method. On real placement shifts, including
shifts to other body locations, we estimate the accuracy of the model after a correction that selects among
all signed permutations of the axes, that is, all reorderings of the three axes combined with sign flips. We call a
shift recovered when the correction brings its accuracy close to the accuracy without the shift. The estimate after the correction
separates the recovered shifts from the others far better than the same estimate computed before the correction. REFIT can therefore keep the corrections that
work and abstain on most of the others (Section~\ref{sec:exp:verify}).
Our main contributions are threefold:
\begin{itemize}
  \item We propose REFIT, a label-free and gradient-free input calibration for frozen wearable models. It undoes
    placement shifts expressed as transforms of the sensor axes, from side switches to the re-attachment of each unit,
    and completes the correction by re-estimating the model's normalization statistics
    (Section~\ref{sec:method:correction}).
  \item We formulate the identification of placement shifts as a comparison of how much of the mismatch in signal
    statistics each family of axis transforms removes on held-out windows of the user (Section~\ref{sec:method:identify}).
    We further introduce verification after the correction, in which a label-free accuracy estimate of the corrected model
    decides when REFIT asks the user to re-wear the sensor (Section~\ref{sec:method:verify}).
  \item Our evaluation covers real left/right sensor pairs, real and simulated re-attachment, injected shifts and
    shifts to other body locations. It shows that identification chooses the right correction and that an accuracy
    estimate after the correction tells when to abstain. We also report data efficiency, computational cost and
    guidance for deployment (Sections~\ref{sec:exp} and~\ref{sec:discussion}).
\end{itemize}

\section{Related Work}
\label{sec:related}

\head{Test-time adaptation for wearable sensing.}
Test-time adaptation updates a trained model with unlabeled data from the deployment
domain~\cite{liang2024ttasurvey}. AdaBN re-estimates the statistics of batch normalization~\cite{ioffe2015batchnorm,adabn}, which already
restores much robustness to corruptions~\cite{schneider2020bnadapt}. TENT tunes the affine normalization parameters by
minimizing prediction entropy~\cite{tent}, and EATA and SAR stabilize this objective against unreliable samples and
collapse~\cite{niu2022eata,niu2023sar}. SHOT adapts the feature extractor by information maximization and
pseudo-labels~\cite{shot}. T3A replaces the classifier by prototypes of confident target samples~\cite{t3a}, and LAME
refines only the output probabilities~\cite{lame}. NOTE makes normalization-based adaptation robust to temporally
correlated test streams, such as those that wearable and smartphone sensors produce~\cite{gong2022note}. For wearable
sensing, OFTTA combines test-time normalization with a prototype classifier to adapt to new users~\cite{oftta}.
SF-Adapter adds a light adapter to a frozen activity model and trains only the adapter~\cite{kang2023sfadapter}.
MetaSense instead adapts to a new user and device from one or two labeled samples per
class~\cite{gong2019metasense}. A systematic study of unsupervised domain adaptation covers differences
in how devices are worn~\cite{chang2020systematic}. Time-series methods adapt without access to the source
data~\cite{mapu}, and domain adaptation for time series is benchmarked in AdaTime~\cite{ragab2023adatime}. PI-TTA
constrains adaptation with gravity, temporal and spectral consistency under sensor rotation and changes of
body location~\cite{li2026pitta}. These methods treat the shift as a change of distribution and adapt statistics, features,
prototypes or adapters. A turned sensor, however, reports each motion on other axes, and none of these updates is designed
to move gravity back to its axis. Input adaptation changes the test input instead of the model, for example by projecting
corrupted images toward the source domain with a diffusion model~\cite{gao2023dda}, but its generative mapping is not a
transform of the sensor axes. In contrast, REFIT estimates the axis transform itself and undoes it in front of the
unchanged model. Normalization re-estimation is only the last step of this correction.

\head{Sensor placement, orientation and device heterogeneity.}
Displacement, placement and orientation of sensors are long-known sources of error in wearable activity
recognition~\cite{kunze2008displacement,kunze2014placement,banos2014displacement}. REALDISP was recorded to study them,
with the same sensors placed once by the experimenters and once partly by the subjects themselves~\cite{realdisp}.
Early work recognized at which body location a sensor is worn~\cite{kunze2005where} and re-calibrated the decision
boundaries of a classifier without labels when sensors were displaced~\cite{forster2009unsupervised}. Later work
extended unsupervised adaptation to sensors that were rotated as well as displaced~\cite{chavarriaga2013unsupervised}
and removed the difference between the two wrists by subspace alignment~\cite{twaites2019wrist}. Gravity has long served
to recover orientation. Averaging the accelerometer gives the vertical direction however the sensor is
mounted~\cite{mizell2003using}, and walking reveals the heading of a phone in a pocket~\cite{kunze2009which}. Other work
removes orientation from the input with orientation-invariant transforms~\cite{yurtman2017orientinv,yurtman2018orient}
or canonical frames derived from gravity and principal axes~\cite{henpraserttae2011orientation}. Others use a transform
to a reference frame in front of a trained accelerometer model~\cite{thiemjarus2010device} or orientation estimates from predefined calibration activities~\cite{friedman2012orientation} or from the repetitive
motion of walking~\cite{wu2013selfcal}. Robustness can also be built into the model
through position-aware recognition~\cite{sztyler2016onbody} or transfer between body locations and
datasets~\cite{wang2018crossposition,
morales2016deep,qin2019cross}. Further options are augmentation with random rotations~\cite{um2017augmentation} or
with the physics of IMU sensing~\cite{xu2023practically,hong2024crosshar}, and rotation-invariant networks for units
re-attached in independent orientations~\cite{baek2026trihar}. A related line measures and mitigates the heterogeneity
of devices, such as differences in their sampling and sensor responses~\cite{stisen2015smart,mathur2018using}. These
approaches discard orientation information that the model could use, require the model to be designed or retrained for
the shift, or rely on a calibration routine or a specific activity. In contrast, REFIT keeps the full signal, works in front of
a model trained without any such provision, and uses only the unlabeled data that the user produces anyway.

\head{Symmetry, canonicalization and mirrored inertial data.}
Group-equivariant networks build a known symmetry into the architecture~\cite{cohen2016group,weiler2019e2}. EqNIO
applies the transformation rules of accelerometers and gyroscopes under rotations and reflections to learn an
equivariant inertial odometry network~\cite{eqnio}. Canonicalization instead maps each input to a standard pose before an
unchanged model. The pose comes from a learned canonicalization function~\cite{kaba2023canonicalization,
mondal2023equivariant} or from a search over inverse transforms at test time, which scores the candidates with the
model's energy or an out-of-distribution score~\cite{schmidt2024its,lindner2026zerottc}. Foundation models can also judge
which candidate looks most typical~\cite{singhal2025focal}. CASTER transports feature
statistics in closed form and uses the residual of the transport to decide when it is unsafe~\cite{khazem2026caster}.
For wrist IMUs, the left/right mirror is established in dietary monitoring, where it is called hand
mirroring~\cite{kyritsis2021eating}. Kyritsis et al.\ map left-wrist smartwatch
recordings onto the right wrist by negating one accelerometer axis and two gyroscope axes, and apply the mirrored data to
a trained network at inference~\cite{kyritsis2021eating}. OREBA publishes the recordings of left-handed subjects
mirrored in the same way~\cite{rouast2020oreba}, and Heydarian et al.\ study this mirroring as a preprocessing step for
intake-gesture detection~\cite{heydarian2020intake}. Silent Impact uses the same mirror to align left-handed and
right-handed tennis players~\cite{park2024silentimpact}. Wang et al.\ apply the mirror to a pretrained eating-gesture
model on external data and choose among mirrored and rotated candidates by the number of samples the model predicts as
eating gestures~\cite{wang2026eating}. Negating every channel of a window, in contrast, is used as a transformation for
self-supervised learning~\cite{saeed2019selfsup,tang2020simclrhar}. Canonicalization methods address images or point
clouds, score each sample with a model, take the kind of transform as given, and search continuous transforms or train a
canonicalizer. The inertial mirror, in turn, assumes a known axis convention, at most up to a rotation about the axis
normal to the sensor face~\cite{wang2026eating}. REFIT
instead fits discrete axis transforms to the raw-signal moments of each sensor unit. Before the model is
queried, it names the shift by how much of the mismatch each family removes on held-out windows. It treats the mirror as
one member of the reflection family, so that a mirror can be found without knowing the axis convention of the device.

\head{Shift diagnosis, accuracy estimation and abstention.}
Work on dataset shift detects a shift~\cite{rabanser2019failing}, quantifies a label shift and corrects the
model~\cite{lipton2018bbse}, and characterizes
which type of shift occurred~\cite{rabanser2019failing,polo2023diagnostics}. A separate line estimates the accuracy of a
model on unlabeled data. The difference of confidences (DoC) and ATC derive
the estimate from the model's confidence~\cite{guillory2021doc,garg2022atc}. ATC, for example, learns a confidence
threshold on labeled source data and predicts the accuracy as the share of unlabeled samples above
it~\cite{garg2022atc}. Other estimates use the agreement
between several models~\cite{baek2022agreement}, self-trained ensembles that also flag misclassified
inputs~\cite{chen2021detecting}, feature distances~\cite{deng2021labels}, or the disagreement of predictions under test-time
adaptation~\cite{lee2024aetta}. In mobile sensing, DAPPER estimates the accuracy of a model adapted to a new user from the
mutual information between its inputs and outputs~\cite{gong2023dapper}. D3M decides without labels whether a shift has
harmed a deployed model~\cite{nguyen2025reliably}. Risk monitoring tracks a model during test-time adaptation and raises
an alert when its performance degrades~\cite{schirmer2025monitoring}. Selective classification rejects individual inputs to keep the
risk of the remaining predictions below a chosen level~\cite{geifman2017selective}. These methods detect or characterize a
shift, assess a given model under it, raise an alarm for retraining, or reject single samples. In contrast, REFIT names a shift by the family of axis
transforms that removes it, and it applies an accuracy estimate after the axes have been corrected, where the estimate
becomes far more reliable. It uses the estimate for each
user and model to decide whether to keep the correction or to ask the user to re-wear the sensor.

\section{Method}
\label{sec:method}

\begin{figure}[t]
\centering
\resizebox{\linewidth}{!}{\begin{tikzpicture}[x=1mm, y=1mm,
  font=\sffamily\fontsize{8.5}{10}\selectfont,
  >={Stealth[length=2.4mm, width=1.8mm]},
  arr/.style={->, line width=0.7pt, black!70},
  harr/.style={->, line width=0.9pt, #1},
  panel/.style={draw=rfDepLn!75!black, fill=white, rounded corners=2pt, minimum height=36mm, inner sep=0pt,
    line width=0.5pt},
  ttl/.style={font=\sffamily\bfseries\fontsize{10.5}{11.5}\selectfont, text=black!85, inner sep=0.5pt},
  net/.style={draw=none, minimum width=19.8mm, minimum height=23.5mm, inner sep=0pt},
  win/.style={draw=none, minimum width=15mm, minimum height=12mm, inner sep=0pt},
  store/.style={draw=rfStore, fill=white, rounded corners=2pt, align=center, inner sep=1.2pt, minimum height=9.5mm,
    line width=0.5pt},
  lbl/.style={align=center, inner sep=0.5pt},
  chip/.style={anchor=west, inner sep=0pt},
  lane/.style={rounded corners=5pt, line width=0.6pt},
  tab/.style={font=\sffamily\fontsize{10}{11}\selectfont, inner xsep=2.5pt, inner ysep=1pt, rounded corners=2pt},
  badge/.style={circle, fill=white, line width=0.5pt, inner sep=0.6pt, font=\fontsize{7.5}{7.5}\selectfont},
]
\definecolor{rfAcc}{HTML}{E8833A}\definecolor{rfGyr}{HTML}{2F6FB3}\definecolor{rfNet}{HTML}{4A86C8}%
\definecolor{rfHead}{HTML}{8E6CC0}\definecolor{rfKey}{HTML}{D35400}\definecolor{rfTrainBg}{HTML}{F8F3EA}%
\definecolor{rfTrainLn}{HTML}{D9CBB4}\definecolor{rfDepBg}{HTML}{EEF4FA}\definecolor{rfDepLn}{HTML}{B9CCE2}%
\definecolor{rfNavy}{HTML}{1F3A5F}\definecolor{rfRed}{HTML}{C0392B}\definecolor{rfOk}{HTML}{2E8B57}%
\definecolor{rfStore}{HTML}{9C7A3C}\definecolor{rfIce}{HTML}{1E88C8}%
\def\rfcube#1#2#3#4#5{%
  \fill[#5!80!black, draw=black!55, line width=0.35pt] (#1+#3,#2) -- ++(2,1.5) -- ++(0,#4) -- (#1+#3,#2+#4) -- cycle;
  \fill[#5!40, draw=black!55, line width=0.35pt] (#1,#2+#4) -- ++(2,1.5) -- ++(#3,0) -- (#1+#3,#2+#4) -- cycle;
  \fill[#5!65, draw=black!55, line width=0.35pt] (#1,#2) rectangle ++(#3,#4);}%
\def\rfnet{%
  \rfcube{-9.9}{-11.75}{3.2}{22}{rfNet}\rfcube{-4.7}{-9.75}{3.2}{18}{rfNet}%
  \rfcube{0.5}{-7.75}{3.2}{14}{rfNet}\rfcube{5.7}{-4.25}{2.2}{7}{rfHead}}%
\def\rfwin#1{%
  \foreach \o in {3,1.5,0}{%
    \fill[white, draw=black!50, line width=0.35pt, rounded corners=0.5pt] (-7.5+\o,-6+\o) rectangle ++(12,9);}%
  \draw[rfAcc, line width=0.65pt] plot[domain=0:10, samples=50, smooth] ({-6.5+\x},{-0.4+1.0*sin(\x*100)*cos(\x*23)});
  \draw[rfGyr, line width=0.65pt] plot[domain=0:10, samples=50, smooth] ({-6.5+\x},{-3.6+0.9*cos(\x*140)});
  \ifnum#1=1
    \foreach \o/\c/\i in {3/rfOk/\faBiking,1.5/rfAcc/\faRunning,0/rfRed/\faWalking}{%
      \node[fill=\c, text=white, rounded corners=0.6pt, inner sep=0pt, minimum width=3.4mm, minimum height=2.6mm,
        font=\fontsize{5.5}{5.5}\selectfont] at (2.8+\o,1.5+\o) {\i};}%
  \fi}%
\def\rfrow{45}%
\node[win] (uw) at (0,0) {};
\node[panel, minimum width=34mm, right=6mm of uw] (rec) {};
\node[panel, minimum width=25mm, right=6mm of rec] (fix) {};
\node[net, right=6mm of fix] (fz) {};
\node[panel, minimum width=24mm, right=7mm of fz] (tst) {};
\node[lbl, right=6.5mm of tst, yshift=7.5mm] (act) {{\fontsize{11}{11}\selectfont\textcolor{rfOk}{\faRunning}}\\[1.5pt]Activity};
\node[lbl, right=6.5mm of tst, yshift=-7.5mm] (rw) {{\fontsize{11}{11}\selectfont\textcolor{rfRed}{\faIcon{undo-alt}}}\\[1.5pt]Ask to\\Re-Wear};
\node[net] (ft) at (fz |- 0,\rfrow) {};
\node[win] (lw) at ($(uw |- 0,\rfrow)+(0,-4)$) {};
\node[store, minimum width=30mm] (mom) at ($(rec |- 0,\rfrow)+(0,-8)$)
  {\textbf{\textcolor{rfStore}{\faDatabase}\ Moments}\\[1pt]$\mu_u,\ \Sigma^{a}_u,\ \Sigma^{\omega}_u,\ \bar{m}$};
\node[store, minimum width=19mm] (tau) at ($(tst |- 0,\rfrow)+(0,-8)$)
  {\textbf{\textcolor{rfStore}{\faDatabase}\ ATC}\\[1pt]\textbf{Threshold} $\tau$};
\node[draw=rfRed, fill=rfRed!8, rounded corners=2pt, minimum width=14mm, minimum height=7mm,
  font=\sffamily\bfseries\fontsize{8.5}{10}\selectfont, line width=0.5pt] (loss) at ($(tst |- 0,\rfrow)+(0,8)$) {Loss};
\coordinate (frL) at ($(uw.west)+(-5.2,0)$);
\coordinate (frR) at ($(rw.east)+(3,0)$);
\begin{scope}[on background layer]
  \fill[rfTrainBg, draw=rfTrainLn, lane] (frL |- 0,\rfrow-16) rectangle (frR |- 0,\rfrow+18.5);
  \fill[rfDepBg, draw=rfDepLn, lane] (frL |- 0,-22.5) rectangle (frR |- 0,21.5);
\end{scope}
\begin{scope}[shift={(uw.center)}] \rfwin{0} \end{scope}
\begin{scope}[shift={(lw.center)}] \rfwin{1} \end{scope}
\begin{scope}[shift={(fz.center)}] \rfnet \end{scope}
\begin{scope}[shift={(ft.center)}] \rfnet \end{scope}
\node[lbl, anchor=south] at ($(uw.north)+(0,1)$) {User's Windows};
\node[lbl, anchor=north, text=black!65] at ($(uw.south)+(0,-1)$) {Unlabeled};
\node[lbl, anchor=south] at ($(lw.north)+(0,1)$) {Labeled Windows};
\node[lbl, anchor=north, text=black!65] at ($(lw.south)+(0,-1)$) {Training Users};
\node[badge, draw=rfIce, text=rfIce] at ($(fz.center)+(-8.3,12.4)$) {\faSnowflake};
\node[badge, draw=rfAcc, text=rfAcc] at ($(ft.center)+(-8.3,12.4)$) {\faFire};
\node[lbl, anchor=north, font=\fontsize{9.5}{9.5}\selectfont] at ($(fz.center)+(-8.3,-12.3)$) {$f$};
\node[lbl, anchor=east] at ($(ft.west)+(-1.5,6.5)$) {Classifier $f$};
\node[ttl, anchor=north] at ($(rec.north)+(0,-1.5)$) {\textcolor{rfKey}{RE}COGNIZE};
\node[anchor=north, text=black!70, inner sep=0.5pt] at ($(rec.north)+(0,-5.6)$) {Fit Each Family};
\foreach \name/\len [count=\i] in {Rotation/4.5, Reflection/10.5, Permutation/6.5, Sign Flip/5.5, Gain/2.6, Offset/1.6}{
  \pgfmathsetmacro{\yy}{9.95-\i*3.35}
  \ifnum\i=2
    \node[chip, font=\sffamily\bfseries\fontsize{8.5}{10}\selectfont, text=rfHead!85!black] at ($(rec.center)+(-14.8,\yy)$) {\name};
    \fill[rfHead] ($(rec.center)+(2.4,\yy-1.0)$) rectangle ++(\len,2.0);
    \node[anchor=west, inner sep=0pt, text=rfHead, font=\fontsize{7.5}{7.5}\selectfont] at ($(rec.center)+(2.4+\len+0.7,\yy)$) {\faCheck};
  \else
    \node[chip, text=black!80] at ($(rec.center)+(-14.8,\yy)$) {\name};
    \fill[black!25] ($(rec.center)+(2.4,\yy-1.0)$) rectangle ++(\len,2.0);
  \fi
}
\draw[black!45, line width=0.4pt] ($(rec.center)+(2.4,8.3)$) -- ($(rec.center)+(2.4,-11.9)$);
\node[anchor=south, inner sep=0.5pt] at ($(rec.south)+(0,1.3)$) {Largest $G_{\mathcal{F}}$, or None};
\node[ttl, anchor=north] at ($(fix.north)+(0,-1.5)$) {\textcolor{rfKey}{FI}X};
\begin{scope}[shift={($(fix.center)+(-6.4,2.2)$)}, line width=0.85pt, ->, >={Stealth[length=1.6mm, width=1.4mm]}]
  \draw[black!45] (0,0) -- (-5,0); \draw[black!45] (0,0) -- (0,5); \draw[black!45] (0,0) -- (2.5,-2.1);
\end{scope}
\begin{scope}[shift={($(fix.center)+(6.6,2.2)$)}, line width=0.85pt, ->, >={Stealth[length=1.6mm, width=1.4mm]}]
  \draw[rfOk] (0,0) -- (5,0); \draw[rfOk] (0,0) -- (0,5); \draw[rfOk] (0,0) -- (-2.5,-2.1);
\end{scope}
\draw[->, line width=0.6pt, black!70] ($(fix.center)+(-3.0,3.4)$) -- ($(fix.center)+(3.0,3.4)$)
  node[midway, above=0.6mm, inner sep=0pt] {$\hat{Q}_{u,\mathcal{F}}$};
\node[align=center, anchor=north, inner sep=0.5pt] at ($(fix.center)+(0,-4.2)$)
  {\textcolor{rfOk}{\faSync}\ Re-Estimate\\Normalization\\Statistics};
\node[ttl, anchor=north] at ($(tst.north)+(0,-1.5)$) {\textcolor{rfKey}{T}EST};
\foreach \h [count=\i] in {8.4,6.2,9.0,3.2,7.6,7.0,2.6,8.6}{
  \pgfmathsetmacro{\xx}{-9.0+\i*1.9}
  \pgfmathtruncatemacro{\ok}{\h>5.0}
  \ifnum\ok=1 \fill[rfOk!75] ($(tst.center)+(\xx,-2.4)$) rectangle ++(1.3,\h);
  \else \fill[black!25] ($(tst.center)+(\xx,-2.4)$) rectangle ++(1.3,\h); \fi}
\draw[black!45, line width=0.4pt] ($(tst.center)+(-8.0,-2.4)$) -- ($(tst.center)+(8.4,-2.4)$);
\draw[rfRed, dashed, line width=0.6pt] ($(tst.center)+(-8.0,2.6)$) -- ($(tst.center)+(8.4,2.6)$)
  node[right=0.3mm, inner sep=0pt, text=rfRed] {$\tau$};
\node[anchor=north, align=center, inner sep=0.5pt] at ($(tst.center)+(0,-4.0)$) {ATC Estimate\\$\geq$ Required\\Accuracy};
\draw[arr] (uw.east) -- (rec.west);
\draw[arr] (rec.east) -- (fix.west);
\draw[arr] (fix.east) -- (fz.west);
\draw[arr] (fz.east) -- (tst.west);
\draw[arr] (tst.east |- act) -- (act.west) node[midway, above=0.3mm, text=rfOk] {\faCheck};
\draw[arr] (tst.east |- rw) -- (rw.west) node[midway, above=0.3mm, text=rfRed] {\faTimes};
\coordinate (jl) at ($(lw.east)+(2.5,0)$);
\draw[line width=0.7pt, black!70] (lw.east) -- (jl);
\draw[arr] (jl) |- (ft.west);
\draw[arr] (jl) |- (mom.west);
\coordinate (jr) at ($(ft.east)+(2.5,0)$);
\draw[line width=0.7pt, black!70] (ft.east) -- (jr);
\draw[arr] (jr) |- (loss.west);
\draw[arr] (jr) |- (tau.west);
\draw[->, rfRed, line width=0.7pt, dashed] (loss.north) -- ++(0,3.0) -| ($(ft.center)+(3.1,8.6)$)
  node[pos=0.25, above=0.3mm, inner sep=0.5pt] {Backprop};
\draw[harr=rfStore] (mom.south) -- (rec.north);
\draw[harr=rfStore] (tau.south) -- (tst.north);
\draw[harr=rfIce] ($(ft.center)+(-2.1,-9.75)$) -- ($(fz.center)+(-2.1,9.3)$)
  node[pos=0.45, left=1mm, font=\sffamily\bfseries\fontsize{9}{10}\selectfont, text=rfIce] {Frozen \faSnowflake};
\node[tab, fill=rfTrainBg, anchor=west] at ($(frL |- 0,\rfrow+18.5)+(4,0)$) {\textbf{TRAINING}\ \ \textcolor{black!65}{Offline, Once}};
\node[tab, fill=rfDepBg, anchor=west] at ($(frL |- 0,-22.5)+(4,0)$) {\textbf{DEPLOYMENT WITH REFIT}\ \ \textcolor{black!65}{Per User}};
\node[fill=rfNavy, text=white, rounded corners=4pt, inner xsep=4pt, inner ysep=1.9pt, anchor=east,
  font=\sffamily\bfseries\fontsize{9}{10}\selectfont] at ($(frR |- 0,-22.5)+(-4,0)$)
  {No Labels\ \ $\cdot$\ \ No Gradient Steps\ \ $\cdot$\ \ Weights Unchanged};
\end{tikzpicture}%
}
\caption{Overview of REFIT. The classifier $f$ is trained once and then frozen. For each user, REFIT recognizes the
family of the placement shift (Section~\ref{sec:method:identify}), fixes it (Section~\ref{sec:method:correction}), and
tests the fixed model before its predictions are used (Section~\ref{sec:method:verify}).}
\label{fig:overview}
\end{figure}
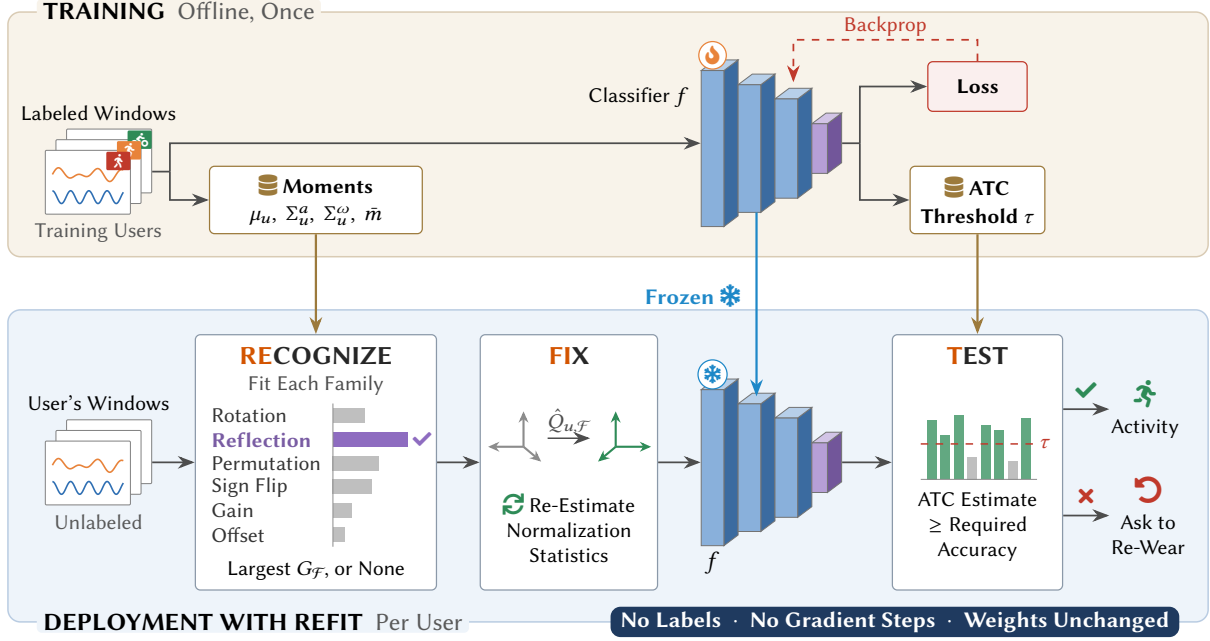

We present REFIT, a label-free input calibration that is placed in front of a frozen activity-recognition model
(Figure~\ref{fig:overview}). REFIT recognizes the kind of placement shift, fixes it, and tests the fixed model before its
predictions are used. Section~\ref{sec:method:setting} defines the setting and the families of axis transforms.
Section~\ref{sec:method:identify} recognizes the family of the shift (identification), Section~\ref{sec:method:correction}
fixes the shift (correction), and Section~\ref{sec:method:verify} tests the fixed model and decides when REFIT abstains
(verification).

\subsection{Setting and Transform Families}
\label{sec:method:setting}

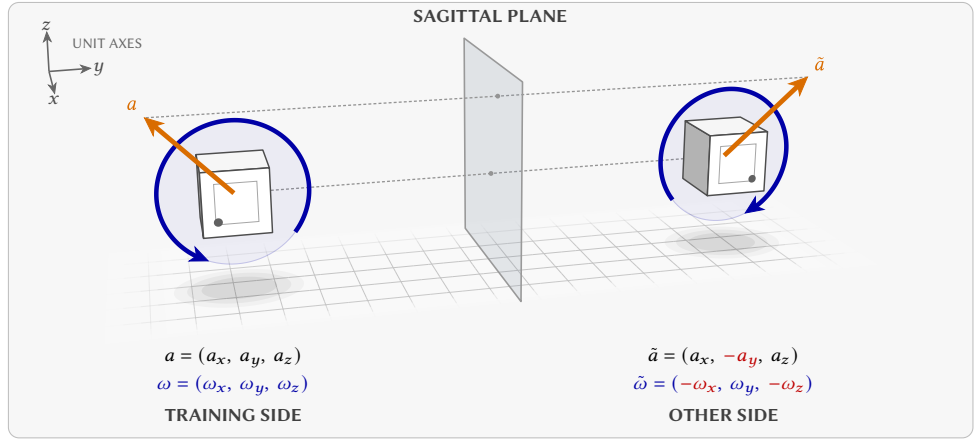
\begin{figure}[t]
\centering
\begin{tikzpicture}[x=1mm, y=1mm,
  font=\sffamily\fontsize{7}{8}\selectfont,
  ttl/.style={font=\sffamily\fontsize{7}{8}\selectfont\bfseries, text=black!75},
  axl/.style={text=black!70, inner sep=0.6pt},
]
\newcommand{\flip}[1]{\textcolor{red!75!black}{#1}}
\useasboundingbox (0,0) rectangle (128,58);
\fill[black!3, draw=black!25, rounded corners=4pt] (0.2,0.2) rectangle (127.8,57.8);
\node[inner sep=0pt, anchor=south west] at (0,0)
  {\animategraphics[loop, width=128mm]{14}{figures/fig_mirror_anim/frame_}{00}{47}};
\node[ttl, anchor=south] at (63.97,53.9) {SAGITTAL PLANE};
\node[axl, anchor=north] at (6.24,45.71) {$x$};
\node[axl, anchor=west] at (10.98,49.02) {$y$};
\node[axl, anchor=south] at (5.20,53.81) {$z$};
\node[text=black!60, anchor=west, font=\sffamily\fontsize{5.5}{6.5}\selectfont] at (7.4,52.4) {UNIT AXES};
\node[align=center, anchor=north] at (29.96,13.4)
  {$a = (a_x,\ a_y,\ a_z)$\\[2pt]\textcolor{blue!65!black}{$\omega = (\omega_x,\ \omega_y,\ \omega_z)$}};
\node[align=center, anchor=north] at (94.82,13.4)
  {$\tilde{a} = (a_x,\ \flip{-a_y},\ a_z)$\\[2pt]\textcolor{blue!65!black}{$\tilde{\omega} = (\flip{-\omega_x},\ \omega_y,\ \flip{-\omega_z})$}};
\node[ttl] at (29.96,3.3) {TRAINING SIDE};
\node[ttl] at (94.82,3.3) {OTHER SIDE};
\end{tikzpicture}
\caption{A side switch mirrors a unit through the body's plane of symmetry (sagittal plane), here normal to its $y$
axis. The acceleration $a$ is mirrored like an ordinary arrow, while a rotation $\omega$ about an axis within the plane
reverses its sense. The gyroscope therefore flips the two axes that the accelerometer keeps (flipped signs in red). Readers can click and play
the animation in this figure using \textcolor{red}{Adobe Acrobat}.}
\label{fig:mirror}
\end{figure}

A classifier $f$ is trained on the labeled windows of a group of training users, whose sensors are worn in the training
placement, and is then frozen. At deployment, it receives the unlabeled windows of one user. A window contains $U$ sensor
units, and unit $u$ provides a tri-axial accelerometer signal $a_u(t)$ and a tri-axial gyroscope signal $\omega_u(t)$.
We model the common placement shifts, side switches and re-attachments, as an orthogonal transform $A_u$ of the axes of
each unit,
\begin{equation}
  \tilde{a}_u(t) = A_u\, a_u(t), \qquad \tilde{\omega}_u(t) = \det(A_u)\, A_u\, \omega_u(t), \qquad A_u \in O(3),
  \label{eq:shift}
\end{equation}
where the tilde marks the signals that the model receives after the shift. The accelerometer measures an ordinary, or
polar, vector and turns with $A_u$. The gyroscope measures an angular velocity, which is an axial vector and takes the
extra factor $\det(A_u)$. Under a rotation ($\det A_u = +1$) both signals turn alike, and under a reflection
($\det A_u = -1$) the gyroscope changes sign relative to the accelerometer (Figure~\ref{fig:mirror}).

\head{Transform families.}
A side switch places a unit at the mirror image of its training position, so its transform is a reflection. If the mirror
plane is normal to axis $j$ of the unit, Eq.~\eqref{eq:shift} flips axis $j$ of the accelerometer and the other two axes
of the gyroscope. For a mirror plane normal to the $y$ axis and channels ordered
$(a_x, a_y, a_z, \omega_x, \omega_y, \omega_z)$, the transform is the fixed sign pattern
\begin{equation}
  s = (+1,\,-1,\,+1,\,-1,\,+1,\,-1),
  \label{eq:mirror}
\end{equation}
applied channel by channel. A re-attachment keeps the body location of a unit but turns its axes. A strap-mounted sensor unit is
a box that can be put back in one of 24 axis-aligned orientations: any of its six faces against the body, turned in
one of four directions. These are the $3 \times 3$
signed permutation matrices with determinant $+1$,
$\mathcal{Q}_{24} = \{Q : Q_{ij} \in \{0, \pm 1\},\, Q^\top Q = I,\, \det Q = +1\}$. For example, a phone put into
a trouser pocket turned by 180 degrees about its $y$ axis reports every motion with the signs of its $x$ and $z$ axes
reversed. Its transform is $A_u = \mathrm{diag}(-1, 1, -1)$ for the accelerometer and the gyroscope alike, since
$\det A_u = +1$. The 24 rotations together with their mirror images form the 48 signed permutations
$\mathcal{Q}_{48}$, which include the reflection across every plane spanned by two
axes of the unit. For identification, REFIT considers six families: rotations, reflections, permutations of the six
channels, per-channel sign flips, and a gain and an offset per channel. The rotations are the 24 members of
$\mathcal{Q}_{24}$. The reflections are the 24 signed permutations with determinant $-1$,
$\mathcal{Q}_{48} \setminus \mathcal{Q}_{24}$, each a mirror image possibly combined with a rotation, including the three
axis reflections. The last four families change
the raw signal without turning the unit, as differences in axis order, polarity, sensitivity or zero level between devices do.
They let identification tell a placement shift from such changes.

\head{Moment matching.}
The correction fits a transform by matching simple statistics of the raw signal, in the spirit of moment matching in
domain adaptation~\cite{coral,peng2019moment}. For unit $u$, let $\mu_u$ be the mean
acceleration over all samples of the training windows, which gravity dominates. Let $\Sigma^{a}_u$ and
$\Sigma^{\omega}_u$ be the covariances of the
accelerometer and the gyroscope samples. The same moments of the user's unlabeled windows carry a tilde. REFIT
scores a candidate correction $Q$, which maps the user's axes back onto the training axes, by the moment discrepancy
\begin{equation}
  D_u(Q) = \frac{\lVert Q\tilde{\mu}_u - \mu_u \rVert^2}{\lVert \mu_u \rVert^2}
  + \frac{\lVert Q\tilde{\Sigma}^{a}_u Q^\top - \Sigma^{a}_u \rVert^2}{\lVert \Sigma^{a}_u \rVert^2}
  + \frac{\lVert Q\tilde{\Sigma}^{\omega}_u Q^\top - \Sigma^{\omega}_u \rVert^2}{\lVert \Sigma^{\omega}_u \rVert^2}.
  \label{eq:discrep}
\end{equation}
Here $\lVert \cdot \rVert$ is the Euclidean norm of a vector and the Frobenius norm of a matrix. The gyroscope factor
$\det Q$ cancels in the covariance. For a family $\mathcal{F}$ of rotations or reflections, the fitted member
$\hat{Q}_{u,\mathcal{F}} = \arg\min_{Q \in \mathcal{F}} D_u(Q)$ is applied to the raw windows as
$\tilde{a}_u \mapsto \hat{Q}_{u,\mathcal{F}}\, \tilde{a}_u$ and
$\tilde{\omega}_u \mapsto \det(\hat{Q}_{u,\mathcal{F}})\, \hat{Q}_{u,\mathcal{F}}\, \tilde{\omega}_u$. Each fit needs only moments
of the raw signal, no forward pass of the model and no gradient. A reflection $Q$ and the rotation $-Q$ give the same
covariances, so Eq.~\eqref{eq:discrep} tells them apart only through the mean acceleration. Identification therefore
also uses the correlations between accelerometer and gyroscope axes, which change sign between the two
(Section~\ref{sec:method:identify}).

\subsection{Recognize: Identifying the Shift}
\label{sec:method:identify}

When the kind of shift is unknown, REFIT names it for each unit by how much of the mismatch each family removes
on windows that its fit has not seen. To tell the families apart, identification describes a unit by a richer vector
$m$ of signal statistics with $K$ components. Per channel, it contains the mean, the logarithm of the standard deviation, the spectral
centroid and the autocorrelation at two lags. It also contains the correlations within the accelerometer and within the
gyroscope, and the nine correlations between accelerometer and gyroscope axes, which tell a reflection from
the corresponding rotation (Section~\ref{sec:method:setting}). A unit with six channels thus has $K = 45$ components. Let $\bar{m}$ be the mean of this vector over
the training users. Let $\sigma_k$ be the larger of two standard deviations of component $k$, one across the
training users and one across eight random batches of 200 training windows. The discrepancy of identification is
\begin{equation}
  \Delta(m) = \frac{1}{K} \sum_{k=1}^{K} \left( \frac{m_k - \bar{m}_k}{\sigma_k} \right)^2 .
  \label{eq:iddisc}
\end{equation}
Each family is fitted directly, without gradient steps. Rotations and reflections take the axis-aligned member that
minimizes Eq.~\eqref{eq:iddisc} rather than Eq.~\eqref{eq:discrep}, so that all six families are scored on the same
statistics. Permutations take the one-to-one assignment of the channels to the training channels with
the most similar statistics. Sign flips follow the polarity of each channel's mean and skewness, gains match the standard deviations,
and offsets match the means. REFIT splits the user's windows into two random halves $\mathcal{A}$
and $\mathcal{B}$, fits each family on one half, applies the fitted member unchanged to the other half, and averages over
both directions. The held-out reduction of family $\mathcal{F}$, a two-fold cross-validated score, is
\begin{equation}
  G_{\mathcal{F}} = \frac{1}{2} \left[
    \frac{\Delta^{\mathcal{B}}(I) - \Delta^{\mathcal{B}}(\hat{Q}^{\mathcal{A}}_{\mathcal{F}})}{\Delta^{\mathcal{B}}(I)}
  + \frac{\Delta^{\mathcal{A}}(I) - \Delta^{\mathcal{A}}(\hat{Q}^{\mathcal{B}}_{\mathcal{F}})}{\Delta^{\mathcal{A}}(I)} \right].
  \label{eq:reduction}
\end{equation}
Here $\Delta^{\mathcal{B}}(T)$ is the discrepancy of Eq.~\eqref{eq:iddisc} on half $\mathcal{B}$ after the transform
$T$ is applied, $\hat{Q}^{\mathcal{A}}_{\mathcal{F}}$ is the member of $\mathcal{F}$ fitted on half $\mathcal{A}$, and
$I$ is the identity. For gains and offsets, this member scales or shifts the channels. Dividing by the discrepancy before the fit
puts users with small and large mismatches on one scale. Scoring each fit on the half it has not seen lets families with
different numbers of parameters compete fairly, because parameters that only fit the noise of one half do not carry
over to the other.

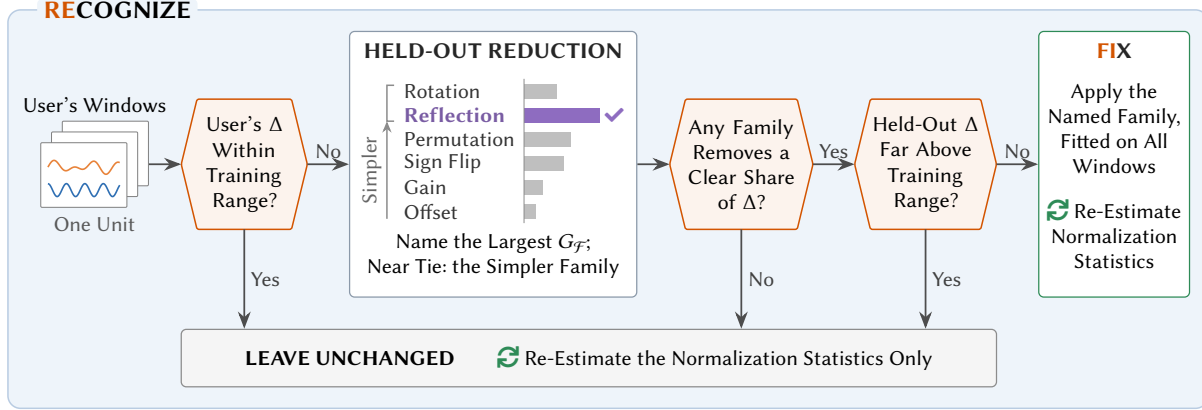
\begin{figure}[t]
\centering
\resizebox{\linewidth}{!}{\begin{tikzpicture}[x=1mm, y=1mm,
  font=\sffamily\fontsize{8.5}{10}\selectfont,
  >={Stealth[length=2.4mm, width=1.8mm]},
  arr/.style={->, line width=0.7pt, black!70},
  panel/.style={draw=rfiDepLn!75!black, fill=white, rounded corners=2pt, inner sep=0pt, line width=0.5pt},
  ask/.style={align=center, inner sep=0.6pt, minimum height=18mm, text width=#1},
  ttl/.style={font=\sffamily\bfseries\fontsize{9}{10}\selectfont, text=black!85, inner sep=0.5pt},
  yn/.style={inner sep=0.6pt, text=black!75},
  win/.style={draw=none, minimum width=15mm, minimum height=12mm, inner sep=0pt},
  lbl/.style={align=center, inner sep=0.5pt},
  chip/.style={anchor=west, inner sep=0pt},
  tab/.style={font=\sffamily\fontsize{10}{11}\selectfont, inner xsep=2.5pt, inner ysep=1pt, rounded corners=2pt},
]
\definecolor{rfiAcc}{HTML}{E8833A}\definecolor{rfiGyr}{HTML}{2F6FB3}\definecolor{rfiHead}{HTML}{8E6CC0}%
\definecolor{rfiKey}{HTML}{D35400}\definecolor{rfiDepBg}{HTML}{EEF4FA}\definecolor{rfiDepLn}{HTML}{B9CCE2}%
\definecolor{rfiOk}{HTML}{2E8B57}%
\def\rfiwin{%
  \foreach \o in {3,1.5,0}{%
    \fill[white, draw=black!50, line width=0.35pt, rounded corners=0.5pt] (-7.5+\o,-6+\o) rectangle ++(12,9);}%
  \draw[rfiAcc, line width=0.65pt] plot[domain=0:10, samples=50, smooth] ({-6.5+\x},{-0.4+1.0*sin(\x*100)*cos(\x*23)});
  \draw[rfiGyr, line width=0.65pt] plot[domain=0:10, samples=50, smooth] ({-6.5+\x},{-3.6+0.9*cos(\x*140)});}%
\def\rfihp{2.2}%
\def\rfidec#1{%
  \draw[rfiKey, fill=rfiKey!7, line width=0.6pt, rounded corners=0.8pt]
    ($(#1.west)+(-\rfihp,0)$) -- (#1.north west) -- (#1.north east) -- ($(#1.east)+(\rfihp,0)$)
    -- (#1.south east) -- (#1.south west) -- cycle;}%
\node[win] (w) at (0,0) {};
\node[ask=12.6mm, right=6.7mm of w] (d1) {User's $\Delta$\\Within\\Training\\Range?};
\node[panel, minimum width=40mm, minimum height=37mm, right=8.0mm of d1] (fit) {};
\node[ask=15mm, right=6.7mm of fit] (d2) {Any Family\\Removes a\\Clear Share\\of $\Delta$?};
\node[ask=14.8mm, right=10.2mm of d2] (d3) {Held-Out $\Delta$\\Far Above\\Training\\Range?};
\node[panel, draw=rfiOk, minimum width=21mm, minimum height=37mm, right=8.0mm of d3] (fix) {};
\coordinate (exL) at ($(d1.west)+(-\rfihp,0)$);
\coordinate (exR) at ($(d3.east)+(\rfihp,0)$);
\node[draw=black!45, fill=black!4, rounded corners=2pt, line width=0.5pt, inner sep=0pt, minimum height=8mm,
  fit={(exL |- 0,-23) (exR |- 0,-31)}] (ex) {};
\node at (ex.center) {\textbf{LEAVE UNCHANGED}\qquad\textcolor{rfiOk}{\faSync}\ Re-Estimate the Normalization
  Statistics Only};
\coordinate (frL) at ($(w.west)+(-4.5,0)$);
\coordinate (frR) at ($(fix.east)+(2.5,0)$);
\begin{scope}[on background layer]
  \fill[rfiDepBg, draw=rfiDepLn, rounded corners=5pt, line width=0.6pt] (frL |- 0,-34) rectangle (frR |- 0,21.5);
  \rfidec{d1}\rfidec{d2}\rfidec{d3}
\end{scope}
\begin{scope}[shift={(w.center)}] \rfiwin \end{scope}
\node[lbl, anchor=south] at ($(w.north)+(0,1)$) {User's Windows};
\node[lbl, anchor=north, text=black!65] at ($(w.south)+(0,-1)$) {One Unit};
\node[ttl, anchor=north] at ($(fit.north)+(0,-1.5)$) {HELD-OUT REDUCTION};
\begin{scope}[shift={($(fit.center)+(0.5,0)$)}]
\foreach \name/\len [count=\i] in {Rotation/4.5, Reflection/10.5, Permutation/6.5, Sign Flip/5.5, Gain/2.6, Offset/1.6}{
  \pgfmathsetmacro{\yy}{13.55-\i*3.35}
  \ifnum\i=2
    \node[chip, font=\sffamily\bfseries\fontsize{8.5}{10}\selectfont, text=rfiHead!85!black] at (-13,\yy) {\name};
    \fill[rfiHead] (3.8,\yy-1.0) rectangle ++(\len,2.0);
    \node[anchor=west, inner sep=0pt, text=rfiHead, font=\fontsize{7.5}{7.5}\selectfont] at (3.8+\len+0.7,\yy) {\faCheck};
  \else
    \node[chip, text=black!80] at (-13,\yy) {\name};
    \fill[black!25] (3.8,\yy-1.0) rectangle ++(\len,2.0);
  \fi
}
\draw[black!45, line width=0.4pt] (3.8,11.9) -- (3.8,-8.2);
\draw[->, black!45, line width=0.6pt, >={Stealth[length=1.6mm, width=1.3mm]}] (-15.3,-7.6) -- (-15.3,6.0);
\draw[black!45, line width=0.6pt] (-14.3,11.0) -- ++(-1.0,0) -- (-15.3,6.0) -- ++(1.0,0);
\node[rotate=90, text=black!60, inner sep=0.3pt, anchor=south] at (-15.9,-0.8) {Simpler};
\end{scope}
\node[lbl, anchor=north] at ($(fit.center)+(0,-9.4)$) {Name the Largest $G_{\mathcal{F}}$;\\Near Tie: the Simpler Family};
\node[ttl, anchor=north] at ($(fix.north)+(0,-1.5)$) {\textcolor{rfiKey}{FI}X};
\node[lbl, anchor=north] at ($(fix.north)+(0,-7.0)$) {Apply the\\Named Family,\\Fitted on All\\Windows};
\node[lbl, anchor=north] at ($(fix.north)+(0,-23.2)$) {\textcolor{rfiOk}{\faSync}\ Re-Estimate\\Normalization\\Statistics};
\draw[arr] (w.east) -- ($(d1.west)+(-\rfihp,0)$);
\draw[arr] ($(d1.east)+(\rfihp,0)$) -- (fit.west) node[yn, midway, above=0.5mm] {No};
\draw[arr] (fit.east) -- ($(d2.west)+(-\rfihp,0)$);
\draw[arr] ($(d2.east)+(\rfihp,0)$) -- ($(d3.west)+(-\rfihp,0)$) node[yn, midway, above=0.5mm] {Yes};
\draw[arr] ($(d3.east)+(\rfihp,0)$) -- (fix.west) node[yn, midway, above=0.5mm] {No};
\draw[arr] (d1.south) -- (d1.south |- ex.north) node[yn, midway, right=0.6mm] {Yes};
\draw[arr] (d2.south) -- (d2.south |- ex.north) node[yn, midway, right=0.6mm] {No};
\draw[arr] (d3.south) -- (d3.south |- ex.north) node[yn, midway, right=0.6mm] {Yes};
\node[tab, fill=rfiDepBg, anchor=west] at ($(frL |- 0,21.5)+(4,0)$) {\textbf{\textcolor{rfiKey}{RE}COGNIZE}};
\end{tikzpicture}%
}
\caption{Identification for one sensor unit, a close-up of RECOGNIZE in Figure~\ref{fig:overview}. Each of the three checks can send the
input to LEAVE UNCHANGED. Otherwise REFIT fixes it with the named family.}
\label{fig:identify}
\end{figure}

Figure~\ref{fig:identify} summarizes the decision. REFIT leaves the input unchanged when the user's discrepancy is no larger than a high percentile of the
discrepancies between each training user and the remaining ones. Otherwise, it names the family with the largest
held-out reduction. Among families whose reduction lies within a small margin of the largest, it prefers the family
that comes first in a fixed order. In this way, a more flexible family does not absorb a shift that a simpler family explains. The
order runs from rotations and reflections, which share the first rank, through permutations, sign flips and gains to
offsets. REFIT also leaves the input unchanged
when no family removes a clear share of the mismatch, or when the named family leaves a held-out discrepancy far above
the range of the training users. The member of the named family, fitted as above on all of the user's windows, is then
applied, and the normalization statistics are re-estimated as in Section~\ref{sec:method:correction}. An input left
unchanged receives only the re-estimation. The thresholds of these
rules were fixed on FORTH-TRACE~\cite{forthtrace}, a dataset with one unit on each wrist that the identification and
verification studies do not use. Appendix~\ref{app:injected} lists~them.

\subsection{Fix: Correcting the Shift}
\label{sec:method:correction}

REFIT fixes the shift in front of the frozen model. After identification, it applies the member of the named family
fitted in Section~\ref{sec:method:identify}. When the kind of shift is known in advance, it selects the correction
directly, as described next for side switches and re-attachment.

\head{Side switches.}
When a user moves a unit to the other side, REFIT selects the member of $\mathcal{Q}_{48}$ that minimizes
Eq.~\eqref{eq:discrep}. This set contains the mirror of every axis-aligned convention, that is, of every way in which the
axes of a device can be aligned with the body's plane of symmetry. REFIT therefore needs no knowledge of the axis
convention. When the convention is known, the fixed mirror of Eq.~\eqref{eq:mirror} in that convention applies directly,
without parameters or data. We also explored choosing among
the three axis reflections by the lowest mean prediction entropy of the frozen model, which performed on par. We use
moment matching, which needs no forward pass.

\head{Re-attachment.}
When a user takes a sensor off and puts it back, REFIT first restores the training frame of every unit and then
adapts the normalization statistics of the model to the user. For each unit, it selects the rotation that best matches
the training moments,
\begin{equation}
  \hat{Q}_u = \mathop{\arg\min}\nolimits_{Q \in \mathcal{Q}_{24}} D_u(Q),
  \label{eq:select}
\end{equation}
and applies it to the raw window before the per-channel standardization of the model's input. Only rotations are
candidates, because a re-attached unit keeps the handedness of its axes. The identity is one of the candidates, so a unit that was
not moved can keep its frame. Some devices report linear acceleration, from which gravity has already been removed,
and the mean acceleration then no longer points along gravity. For such devices, a variant of the selection detects
the affected units from the share of the accelerometer's second moment carried by its mean. It selects their rotation
from the two covariance terms alone and keeps the identity unless these differ from the training data more than they
do for most training users (Appendix~\ref{app:gravity}).

\head{Normalization re-estimation.}
On the corrected windows, we re-estimate the batch-normalization statistics of the frozen model, as AdaBN does for
uncorrected inputs~\cite{adabn}, and classify every window with them. This absorbs part of the remaining strap
misalignment and the user's own way of moving. The weights of the model are never updated. When the selection keeps the identity for every
unit, REFIT reduces to this re-estimation. For a model without batch normalization, such as one that
normalizes each window on its own, this step is empty. We also explored aligning the penultimate features in the spirit
of CORAL~\cite{coral} instead of re-estimating the normalization statistics, and searching all 48 signed permutations
for re-attachment. Both were less accurate under arbitrary re-attachment, a random rotation of the unit (Section~\ref{sec:exp:ablation}), and we
therefore use the 24 rotations followed by normalization re-estimation.

\subsection{Test: Verifying the Fix}
\label{sec:method:verify}

After the correction, REFIT estimates the accuracy of the corrected model on the user's unlabeled windows with
average thresholded confidence (ATC)~\cite{garg2022atc}. On labeled windows of the training users that were held out
from training, ATC chooses a threshold $\tau$ on the maximum softmax probability. At this threshold, the share of windows
above $\tau$ equals the accuracy of the model on all of these held-out windows. The estimated accuracy for the user is the share of corrected windows
whose maximum softmax probability exceeds $\tau$. The estimate uses the forward pass that classifies the corrected
windows anyway, the single threshold $\tau$ stored with the model, and no labels of the user.

When the estimate falls below the accuracy that the application requires, REFIT abstains: it does not use the
predictions of the corrected model and asks the user to put the sensor back in place. Otherwise, it keeps the
correction and the re-estimated statistics for the windows that follow.

\section{Experiments}
\label{sec:exp}

We first evaluate the correction on side switches and on real and simulated re-attachment, and then the
identification of the shift and the verification of the corrected model. An ablation and a study of data efficiency and
cost close the section. We evaluate the steps of REFIT separately, each with its own classifier, datasets and configuration
(Table~\ref{tab:studies}): the correction when the kind of shift is known, identification when it is not, and
verification after the correction over the 48 signed permutations, which needs no identification. Numbers are
therefore comparable within a study but not across studies.

\subsection{Experimental Setup}
\label{sec:exp:setup}

\begin{table}[t]
\caption{The five studies of Section~\ref{sec:exp}: data, classifier, configuration of REFIT, and where the results are.}
\label{tab:studies}
\scriptsize
\setlength{\tabcolsep}{0pt}
\begin{tabular*}{\linewidth}{@{\extracolsep{\fill}}lllll}
\toprule
Study & Data & Classifier & REFIT & Results \\
\midrule
Side switches & 5 datasets, 62 held-out subjects & one unit, group normalization & selection over $\mathcal{Q}_{48}$ & Table~\ref{tab:mirror} \\
Real re-attachment & REALDISP, 15 subjects & all units, batch normalization & selection over $\mathcal{Q}_{24}$, re-estimation & Table~\ref{tab:realdisp} \\
Simulated re-attachment & 8 datasets, 116 held-out subjects & all units, batch normalization & selection over $\mathcal{Q}_{24}$, re-estimation & Figure~\ref{fig:reattach} \\
Identification & 9 datasets, 133 held-out subjects & one unit, batch normalization & named family among six, re-estimation & Tables~\ref{tab:naming}, \ref{tab:identify} \\
Verification & 6 datasets, 65 held-out subjects & one unit, batch normalization & selection over $\mathcal{Q}_{48}$, re-estimation, ATC & Table~\ref{tab:verify}, Figure~\ref{fig:verify} \\
\bottomrule
\end{tabular*}
\end{table}

\head{Benchmarks.}
Table~\ref{tab:studies} summarizes the five studies and the configuration of REFIT in each.
\begin{itemize}
\item[(1)] \textbf{Side switches.} Five datasets record the same activities with units on both sides of the body.
DSADS~\cite{dsads} has 8 subjects with units on both arms and legs, and REALDISP~\cite{realdisp} has 17 subjects with
units on both forearms, upper arms, calves and thighs. OPPORTUNITY~\cite{opportunity} has 4 subjects with units on both
upper and lower arms. HuGaDB~\cite{hugadb} has 18 subjects with units on both thighs, shins and feet, and
FORTH-TRACE~\cite{forthtrace} has 15 subjects with a unit on each wrist. For every left/right pair and both directions,
the classifier is trained on one side of the training subjects and tested on the other side of the held-out subject.
No public dataset records the same subjects before and after they move a sensor to the other side, so the other side of
held-out subjects serves as the side switch. This gives 322 combinations of left/right pair, direction and held-out
subject from 62 subjects.
\item[(2)] \textbf{Real re-attachment.} REALDISP~\cite{realdisp} records its 17 subjects with nine units on the arms, back
and legs while they perform 33 activities. In one recording the experimenters attach every unit at its predefined
place (ideal placement). In a second recording (self-placement) each subject puts three of the nine units on without
guidance on their exact
placement, and the experimenters attach the other six. We train on the first recording and test on the second, both for
the same subjects and for held-out subjects. In the official files the second recordings of two subjects
contain only zeros on every unit, so we test on the other 15 subjects. REALDISP also contains recordings of three
subjects in which the experimenters deliberately displaced several units (mutual displacement), which enter the verification
study and the accuracy over all test cases of the identification study.
\item[(3)] \textbf{Simulated re-attachment.} We simulate the re-attachment of the units of held-out subjects on DSADS, HuGaDB,
OPPORTUNITY, UCI-HAR~\cite{ucihar}, MHEALTH~\cite{mhealth}, PAMAP2~\cite{pamap2}, MotionSense~\cite{motionsense} and
USC-HAD~\cite{uschad}, with 116 subjects and one to six units per dataset in total. We hold out one subject at a time and
re-attach each of the subject's units in one of two ways. An axis-aligned re-attachment turns the unit into one of the
24 orientations of the box and adds a strap misalignment of up to 20 degrees about a random axis. An arbitrary
re-attachment, in which the unit can land in any orientation, applies a uniformly random rotation. We also test each subject
with no re-attachment.
\item[(4)] \textbf{Identification.} Here the classifier takes a single unit. For every unit of the eight
datasets above and of REALDISP, we train it on that unit of the training subjects and test it on the held-out subject.
This gives 133 held-out subjects. We inject each family of Section~\ref{sec:method:setting} at a
mild and a strong magnitude into the held-out subject's unit (Appendix~\ref{app:injected}). For rotations, the mild
shift is an axis-aligned re-attachment and the strong one a uniformly random rotation. We also present the unchanged
unit, which should be left as it is. Identification also receives the real left/right pairs and the REALDISP
recordings described next, but not the shifts to other body locations. We call each combination of held-out subject,
unit, injected or real condition and source model a test case.
\item[(5)] \textbf{Verification.} With the same single-unit classifiers, we test the six datasets with several units, DSADS, HuGaDB,
MHEALTH, OPPORTUNITY, PAMAP2 and REALDISP, on 2,202 real placement shifts of 65 held-out subjects. Each shift
presents the model of one unit with windows recorded in another placement. Of these shifts, 292 come from the unit on
the other side of the body (real left/right pairs) and 1,748 are shifts to other body locations. On REALDISP, 135 come
from the same unit in the second recording and 27 from recordings in which the experimenters deliberately displaced
several units. The three source models thus give 6,606 cases of real placement shifts. The rules of identification were set on
FORTH-TRACE, which these two studies do not use.
\end{itemize}

\head{Baselines.}
We compare each step of REFIT with label-free methods for the same task.
\begin{itemize}
\item[(1)] \textbf{Adaptation methods.} We compare with the label-free adaptation methods AdaBN~\cite{adabn}, TENT~\cite{tent}, CORAL~\cite{coral},
SHOT~\cite{shot}, T3A~\cite{t3a}, LAME~\cite{lame} and OFTTA~\cite{oftta}, and with the source model, the frozen model
without any calibration. All methods adapt the same frozen source models with the settings of their original papers,
and no setting is tuned on a test subject. CORAL is applied at test time to the penultimate features. The classifier of the side-switch benchmark has no batch normalization, so
AdaBN does not apply there and OFTTA reduces to T3A. CORAL is not evaluated on side switches.
\item[(2)] \textbf{Identification.} We also compare with a confidence selector and with always correcting a rotation, which applies the rotation
family of identification (Section~\ref{sec:method:identify}) to every unit whatever the shift. The confidence
selector applies each family's correction, or none, and keeps the one that gives the frozen model the highest mean
confidence. Test-time canonicalization ranks transformed inputs by a related confidence score~\cite{schmidt2024its},
although the most confident of several augmented inputs is often not the one closest to the training
data~\cite{shanmugam2021tta}. Both baselines are followed by normalization re-estimation. As a reference, we report the accuracy when the true
family is given.
\item[(3)] \textbf{Verification.} We compare ATC with two other label-free estimates. The difference of confidences
(DoC) lowers the accuracy on held-out training windows by the drop in mean confidence~\cite{guillory2021doc}. The
negative mean entropy of the model's predictions on the held-out subject's windows serves as a third estimate. Each
estimate is computed after the correction and, for comparison, with the source model on the uncorrected windows.
\end{itemize}

\head{Implementation.}
For side switches, the classifier is a three-layer one-dimensional convolutional network with group
normalization~\cite{groupnorm} that takes the accelerometer and gyroscope channels of one unit, each window normalized per
channel. We train three source models with different seeds. For re-attachment, the classifier is a four-layer
one-dimensional convolutional network with batch normalization that takes the accelerometer and gyroscope channels of
all units. We train twelve source models with different seeds. For identification and verification, this four-layer network
takes the six channels of one unit, and we train three source models with different seeds.

\head{Metrics.}
We report classification accuracy averaged over the seeds. For side switches, accuracy is averaged over the left/right
pairs of a dataset and then over its held-out subjects. On REALDISP, it is computed over all windows of all
subjects, and on the simulated benchmark it is the mean over subjects. Differences are tested at the subject level
with the exact Wilcoxon signed-rank test~\cite{wilcoxon1945}, with Holm's adjustment~\cite{holm1979} over the
comparisons of each study. For held-out subjects with no re-attachment, REFIT should not lose accuracy. A
non-inferiority test checks that it stays within 0.01 of AdaBN: the mean difference must exceed $-0.01$, and a
one-sided signed-rank test of the differences shifted by 0.01 must be significant. Identification is measured by the share of cases in which the family is named correctly or an unchanged
unit is left as it is. A placement shift counts as recovered when the accuracy after the correction comes within 0.10
of a reference accuracy. This reference is the accuracy that the same model reaches, after normalization
re-estimation, on the same subject's windows from the unit it was trained for. Verification is measured by the area
under the receiver operating characteristic curve (AUROC) of each estimate for separating recovered shifts from the
others. For identification and verification, intervals and one-sided p-values come from a bootstrap over held-out
subjects.

\subsection{Correction of Side Switches}
\label{sec:exp:mirror}

\begin{table}[t]
\caption{Accuracy on the other side of held-out subjects for a model trained on one side. REFIT selects among the 48 signed permutations without knowing the axis convention of the device. OFTTA equals T3A, as the model has no batch normalization. Best in bold, second best underlined.}
\label{tab:mirror}
\scriptsize
\setlength{\tabcolsep}{0pt}
\begin{tabular*}{\linewidth}{@{\extracolsep{\fill}}lcccccc}
\toprule
Method & DSADS & REALDISP & OPPORTUNITY & HuGaDB & FORTH-TRACE & Mean \\
\midrule
Source model & 0.222 & 0.410 & 0.465 & 0.675 & 0.440 & 0.442 \\
TENT & 0.213 & 0.411 & 0.485 & 0.670 & 0.431 & 0.442 \\
SHOT & \underline{0.291} & \underline{0.500} & 0.371 & \underline{0.863} & \underline{0.491} & \underline{0.503} \\
T3A & 0.250 & 0.435 & 0.458 & 0.757 & 0.458 & 0.471 \\
LAME & 0.153 & 0.335 & \underline{0.487} & 0.646 & 0.361 & 0.397 \\
OFTTA & 0.250 & 0.435 & 0.458 & 0.757 & 0.458 & 0.471 \\
\midrule
REFIT (ours) & \textbf{0.681} & \textbf{0.694} & \textbf{0.665} & \textbf{0.900} & \textbf{0.585} & \textbf{0.705} \\
\bottomrule
\end{tabular*}
\end{table}

Table~\ref{tab:mirror} reports the five datasets. The source model is poor on the other side of the body, because the
mirrored motion reaches it with axes of the wrong sign. Its accuracy ranges from 0.675 on HuGaDB down to 0.222 on
DSADS. The adaptation baselines recover little of this loss on average, because they adjust statistics, features or
prototypes but leave the mirrored axes in place. SHOT gains most on HuGaDB but falls below the source model on
OPPORTUNITY. Without knowing the axis convention of the device, REFIT is better than every baseline on every
dataset. Its mean is 0.705, against 0.503 for SHOT, the strongest baseline, and
0.751 for the best transform chosen with labels.
REFIT is more accurate than SHOT for 55 of the 62 held-out subjects (Holm-adjusted $p = 1.21 \times 10^{-12}$).
Against T3A, OFTTA and LAME, it is more accurate for all 62 (Holm-adjusted $p = 1.73 \times 10^{-18}$).

The search over the 48 signed permutations matters because devices differ in their axis convention. The wrist units of
FORTH-TRACE follow a different convention from the units of the other four datasets. There, the fixed mirror of Eq.~\eqref{eq:mirror}, a mirror of the kind used in dietary
monitoring~\cite{kyritsis2021eating} that matches the convention of the other four datasets, flips the wrong axes. Its accuracy drops
to 0.363, below the 0.440 of the source model without any correction.
REFIT instead finds a transform that fits the device and reaches 0.585. In 93.3\% of the cases it picks the
transform that labels would choose, and the mirror of Eq.~\eqref{eq:mirror} in only 3.33\%. Its accuracy comes within
0.00551 of that of the label-chosen transform.
REFIT thus covers every axis convention without being told which one a device uses.

\subsection{Correction of Re-attachment}
\label{sec:exp:reattach}

\begin{table}[t]
\caption{Accuracy on the second REALDISP recording, in which each subject re-attached three of the nine units, for the same and for held-out subjects. Best in bold, second best underlined.}
\label{tab:realdisp}
\scriptsize
\setlength{\tabcolsep}{0pt}
\begin{tabular*}{\linewidth}{@{\extracolsep{\fill}}lccccccccc}
\toprule
Test subjects & Source model & AdaBN & TENT & CORAL & SHOT & T3A & LAME & OFTTA & REFIT (ours) \\
\midrule
Same subjects & 0.584 & 0.830 & 0.832 & 0.733 & \underline{0.842} & 0.616 & 0.463 & 0.839 & \textbf{0.963} \\
Held-out subjects & 0.547 & 0.808 & 0.810 & 0.703 & \underline{0.827} & 0.578 & 0.421 & 0.819 & \textbf{0.955} \\
\bottomrule
\end{tabular*}
\end{table}

\head{Real re-attachment.}
Table~\ref{tab:realdisp} reports REALDISP. REFIT raises the accuracy of the frozen model from 0.547 to 0.955 for
held-out subjects. It is the best method for the same and for held-out subjects and leads the strongest baseline, SHOT, by 0.128 for
held-out subjects and by 0.121 for the same subjects.
It improves on SHOT for 13 of the 15 held-out subjects ($p = 6.10 \times 10^{-4}$).
Normalization and entropy methods recover only part of the loss. A re-attached unit reports its motion in a turned
frame, which statistics of the unrotated input cannot undo. LAME, which refines the predictions of the source model,
even falls below it.
The selection explains where this gain comes from. For 13 of the 15 subjects, it turns at least one of the nine
units, and 22.2\% of all unit selections are a turn rather than the identity. The turned units lie on the arms, on the
legs and on the back. The selection is the same in all twelve seeds. The largest gain occurs for a subject whose right forearm, left upper arm and left calf
units are turned: the accuracy rises from 0.586 with AdaBN to 0.993. For the two subjects whose units are all kept,
REFIT equals AdaBN.

\begin{figure}[t]
\centering
\includegraphics[width=\linewidth]{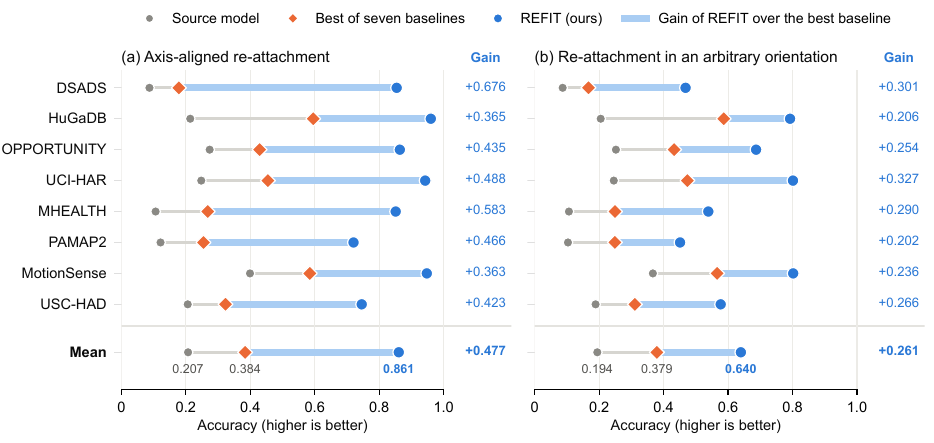}
\caption{Accuracy of held-out subjects under simulated re-attachment on eight datasets. On every dataset REFIT (blue)
is the most accurate. The light-blue bar and the Gain column give its margin over the best of the seven label-free
baselines. In the Mean row, the best baseline is the one with the highest mean. Appendix~\ref{app:baselines} lists
every baseline.}
\label{fig:reattach}
\end{figure}

\head{Simulated re-attachment.}
Figure~\ref{fig:reattach} reports the eight datasets. Under axis-aligned re-attachment REFIT reaches a mean accuracy of
0.861 over the datasets, while the best baseline stays at 0.384. REFIT is better than every baseline for all 116
subjects. It closes 96.7\% of the gap between the source model and the accuracy that normalization re-estimation reaches with no
re-attachment (0.883, Table~\ref{tab:ablation}), and TENT,
the baseline with the highest mean, closes 26.2\%. The selection finds the orientation of the re-attached unit for
92.4\% of the units.
The margin holds on every dataset, from a single phone in a trouser pocket (MotionSense) to six leg units (HuGaDB). It
ranges from 0.363 over the best baseline on MotionSense to 0.676 on DSADS, where all five units are re-attached at
once. On average REFIT comes within 0.0222 of this accuracy with no re-attachment. On
MotionSense it even exceeds that accuracy, because some subjects apparently carried the phone turned from the start and
the selection turns it back.
Under arbitrary re-attachment, where no candidate matches the rotation exactly, REFIT is again the best method on every
dataset. Its mean is 0.640 against 0.379 for the best baseline, and it is better than every baseline for 115 of the 116
subjects. It closes 64.7\% of the gap, against 26.8\% for TENT, and its margin over the best baseline ranges from 0.202
on PAMAP2 to 0.327 on UCI-HAR. The margin is smaller than under axis-aligned re-attachment because the closest of the 24
orientations leaves a residual rotation that normalization re-estimation absorbs only in part
(Appendix~\ref{app:limits}).
On MotionSense and USC-HAD, the two datasets not used to design REFIT, the pre-specified comparisons confirm that
REFIT beats AdaBN and TENT under both kinds of re-attachment. This holds for all 38 subjects (Holm-adjusted
$p = 3.64 \times 10^{-11}$) and in each of the twelve seeds.
When no unit was re-attached, the selection keeps the identity for most subjects and REFIT reduces to normalization
re-estimation. On these two datasets, the pre-specified non-inferiority test confirmed that it stays within 0.01 of
AdaBN in this case. The mean difference is $+0.00412$ (Holm-adjusted $p = 3.67 \times 10^{-9}$). For six MotionSense
subjects who apparently carried the phone the other way round, the selection turns it back by 180 degrees and raises
their accuracy by between 0.0668 and 0.165.

\subsection{Identification of the Shift}
\label{sec:exp:identify}

\begin{table}[t]
\caption{Share of injected shifts for which identification names the family correctly, and share of unchanged units that it leaves as they are (No shift). Best in bold.}
\label{tab:naming}
\scriptsize
\setlength{\tabcolsep}{0pt}
\renewcommand{\arraystretch}{1.0}
\begin{tabular*}{\linewidth}{@{\extracolsep{\fill}}lccccccc}
\toprule
Method & No shift & Rotation & Reflection & Permutation & Sign flip & Gain & Offset \\
\midrule
Confidence selector & 0.347 & 0.373 & 0.662 & 0.424 & \textbf{0.459} & 0.434 & 0.488 \\
REFIT (ours) & \textbf{0.799} & \textbf{0.703} & \textbf{0.772} & \textbf{0.810} & 0.373 & \textbf{0.838} & \textbf{0.972} \\
\bottomrule
\end{tabular*}
\end{table}

\head{Naming the family.}
Table~\ref{tab:naming} reports how often REFIT names the injected family. Over all injected shifts and unchanged units,
REFIT names the family correctly in 0.749 of the
cases, against 0.464 for the confidence selector (one-sided $p < 0.001$). It names the family correctly more often
than the confidence selector for every family except sign flips, and it leaves most unchanged units as they are. The
confidence selector, in contrast, applies a correction to 65.3\% of the unchanged units, because for most of them some
transform raises the confidence of the model. It also often keeps such a correction when the true shift belongs to
another family. Sign flips
are harder to name. A reflection or a rotation also reverses the polarity of axes and often removes as much of the
mismatch, so REFIT prefers these simpler families. A flipped gyroscope, in turn, changes only the gyroscope means, which are
close to zero, and the correlations with the accelerometer, so the input is often left unchanged.
Stronger shifts are named more reliably. For five of the six families, the share named correctly grows from the mild to
the strong magnitude, for example from 0.727 to 0.948 for gains and from 0.699 to 0.920 for permutations. The exception
is the rotation, whose strong form is a uniformly random rotation that falls between the 24 axis-aligned candidates. Its
share drops from 0.865 to 0.542, but the rotation is still the family named most often for these units.
Across the nine datasets, the share named correctly ranges from 0.656 on USC-HAD to 0.847 on MHEALTH
(Appendix~\ref{app:perds}).

\begin{table}[t]
\caption{Accuracy after the correction or adaptation, per family of injected shifts and on real placement shifts, with single-unit classifiers (Table~\ref{tab:studies}); REFIT applies the family that identification names. The real-shift columns therefore differ from Tables~\ref{tab:mirror}, \ref{tab:realdisp} and~\ref{tab:ablation}, which use other classifiers and corrections. Best in bold, second best underlined.}
\label{tab:identify}
\scriptsize
\setlength{\tabcolsep}{0pt}
\renewcommand{\arraystretch}{1.0}
\begin{tabular*}{\linewidth}{@{\extracolsep{\fill}}lccccccccc}
\toprule
& & \multicolumn{6}{c}{Injected shift} & \multicolumn{2}{c}{Real placement shift} \\
\cmidrule(lr){3-8}\cmidrule(lr){9-10}
Method & No shift & Rotation & Reflection & Permutation & Sign flip & Gain & Offset & Left/right pair & Re-attachment \\
\midrule
Source model & 0.843 & 0.196 & 0.380 & 0.183 & 0.406 & 0.398 & 0.337 & 0.566 & 0.646 \\
AdaBN & 0.861 & 0.343 & 0.467 & 0.408 & 0.575 & 0.801 & 0.860 & 0.610 & 0.695 \\
TENT & \underline{0.868} & 0.353 & 0.478 & 0.422 & 0.588 & \underline{0.809} & \textbf{0.868} & 0.621 & 0.701 \\
CORAL & 0.844 & 0.258 & 0.420 & 0.259 & 0.456 & 0.533 & 0.445 & 0.591 & 0.668 \\
SHOT & \textbf{0.880} & 0.306 & 0.459 & 0.372 & 0.572 & 0.767 & 0.777 & 0.626 & 0.736 \\
T3A & 0.863 & 0.207 & 0.396 & 0.193 & 0.428 & 0.439 & 0.368 & 0.585 & 0.667 \\
LAME & 0.825 & 0.188 & 0.365 & 0.174 & 0.381 & 0.379 & 0.319 & 0.542 & 0.608 \\
OFTTA & \underline{0.868} & 0.332 & 0.466 & 0.399 & 0.569 & 0.801 & 0.856 & 0.612 & 0.701 \\
Always rotation & 0.860 & \textbf{0.744} & 0.639 & 0.384 & 0.575 & 0.670 & 0.467 & 0.643 & \textbf{0.809} \\
Confidence selector & 0.855 & 0.562 & \underline{0.747} & \underline{0.577} & \textbf{0.687} & 0.759 & 0.715 & \underline{0.836} & 0.779 \\
\midrule
REFIT (ours) & 0.863 & \underline{0.714} & \textbf{0.838} & \textbf{0.775} & \underline{0.675} & \textbf{0.858} & \underline{0.861} & \textbf{0.841} & \underline{0.802} \\
\bottomrule
\end{tabular*}
\end{table}

\head{Accuracy after the correction.}
Table~\ref{tab:identify} reports the accuracy of the frozen model after the correction. REFIT is the only method that
stays accurate under every kind of shift. Its lowest accuracy over the nine conditions is 0.675, against 0.562 for the
confidence selector and at most 0.384 for every other method. No other method is the most accurate on more than two
conditions or by more than 0.0304.
Over all test cases, REFIT raises the accuracy to 0.794, against 0.696 for the confidence selector, 0.607 for always
correcting a rotation, 0.599 for AdaBN and 0.610 for TENT, the strongest adaptation method. REFIT is more accurate than the
last three overall (Holm-adjusted $p < 0.01$ for each) and on each of the nine datasets.
Over the injected shifts, the unchanged units and the real left/right
pairs, whose family is known, it comes within 0.0272 of the accuracy reached when the true family is given.
The families differ in how much they harm the model and in how much normalization re-estimation alone repairs.
Gains and offsets change only the scale and the level of each channel, which the re-estimated statistics largely absorb: AdaBN
alone reaches 0.801 and 0.860 there. Rotations, reflections and permutations change which axis carries which part of
the motion, and AdaBN stays between 0.343 and 0.467. With the correction of the named family, REFIT reaches 0.714 to
0.838 on these three families and is the most accurate method on reflections, permutations and gains. Always correcting a rotation is more accurate on
rotations, because it assumes the right family from the start, but it falls below AdaBN alone on permutations, gains
and offsets. The confidence selector is slightly more accurate on sign flips, which REFIT names less often. When nothing
was injected, REFIT keeps the accuracy of AdaBN. SHOT, TENT and OFTTA are more accurate there, and TENT also on
offsets. This is because they adapt the weights, the normalization parameters or the classifier of the model to each
subject, which REFIT leaves unchanged.

\head{Real placement shifts.}
On the real left/right pairs, REFIT is the most accurate method, while always correcting a rotation cannot express the
mirror and reaches only 0.643. REFIT names the reflection for 65.0\% of the units and leaves 28.2\% unchanged, 89.1\% of them
because their statistics already lie within the range of the training subjects. On real
re-attachment, it comes within 0.00718 of always correcting a rotation, which assumes the right family there from the
start. It leaves 68.6\% of these units unchanged, names a rotation for 16.5\% and another family for the rest. In
REALDISP each subject attached only three of the nine units, and the experimenters attached the others at their
standard place. Many units are therefore expected to need no correction.

\subsection{Verification of the Correction}
\label{sec:exp:verify}

To measure verification separately from identification, the verification study applies one correction to every
placement shift, with the single-unit classifiers of Section~\ref{sec:exp:setup}. It selects among the 48 signed permutations with Eq.~\eqref{eq:discrep}, a set that contains both
the mirror of a side switch and the rotations of a re-attachment. The normalization statistics are then re-estimated.
On the real placement shifts, this correction recovers 84.2\% of the left/right pairs and 70.9\% of the REALDISP
re-attachments. It recovers only 1.41\% of the shifts to other body locations, in which the model receives the
windows of a unit worn elsewhere. REFIT therefore has to tell these cases apart. The correction raises the mean
accuracy of the left/right pairs from 0.610 with AdaBN alone to 0.815 and that of the re-attachments from 0.695 to
0.754. A shift to another body location alters the motion itself, not only its axes, and the correction raises its
accuracy only from 0.257 to 0.274.

\begin{figure}[t]
\centering
\includegraphics[width=\linewidth]{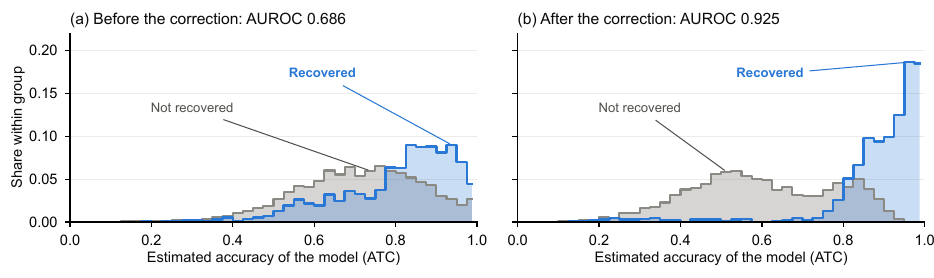}
\caption{ATC estimate for the real placement shifts before and after the correction over the 48 signed permutations,
for the recovered shifts and for those not recovered. After the correction, the estimate separates the two groups.}
\label{fig:verify}
\end{figure}

Table~\ref{tab:verify} reports how well each label-free estimate separates the recovered shifts from the others. After
the correction, all three estimates separate them well. ATC, the estimate that REFIT uses, reaches the highest
AUROC, 0.925. Computed before the correction, the same estimates reach
only 0.686 to 0.716.
Figure~\ref{fig:verify} shows why the correction matters. Before the correction, the estimate describes a model that
still sees turned or mirrored axes, and the recovered shifts and those not recovered receive similar estimates. After
the correction, the recovered shifts receive estimates close to the accuracy of the model on the training subjects.
The shifts to other body locations, which no axis transform undoes, receive low estimates.

\begin{table}[t]
\caption{AUROC of label-free accuracy estimates for separating the real placement shifts that the correction over the 48 signed permutations recovers from those it does not, before and after the correction, with 95\% bootstrap intervals in parentheses. Best in bold, second best underlined.}
\label{tab:verify}
\scriptsize
\setlength{\tabcolsep}{0pt}
\begin{tabular*}{\linewidth}{@{\extracolsep{\fill}}lccc}
\toprule
AUROC & DoC & Negative entropy & ATC (used by REFIT) \\
\midrule
Before correction & \textbf{0.716} (0.695--0.736) & \underline{0.705} (0.684--0.725) & 0.686 (0.663--0.706) \\
After correction & \underline{0.880} (0.822--0.918) & 0.852 (0.794--0.889) & \textbf{0.925} (0.861--0.965) \\
\bottomrule
\end{tabular*}
\end{table}

The separation does not rest on the shifts to other body locations alone. In an exploratory analysis, ATC after the
correction separates the recovered shifts within the left/right pairs with an AUROC of 0.914, against 0.553 before
the correction. Within the REALDISP re-attachments, it reaches 0.913, against 0.672. For these two kinds of shift, the
estimate after the correction also lies close to the true accuracy of the corrected model, with a mean absolute error
of 0.0694 and 0.0869. Before the correction, it misses the accuracy of the uncorrected model by 0.226 and 0.216. REFIT
can therefore ask the user to re-wear the sensor when the estimate is low instead of silently keeping a correction
that does not recover the shift.

A simple rule shows what this means for a user. In an exploratory analysis, we set
the requirement of Section~\ref{sec:method:verify} relative to the model. REFIT keeps the correction whenever the ATC
estimate comes within 0.10 of the model's accuracy on held-out windows of the training subjects, which mirrors the
definition of a recovered shift without using its labels. This rule keeps 85.2\% of the recovered shifts and
abstains on 95.2\% of the others. The kept shifts reach a mean accuracy of 0.831 and the abstained ones 0.275. It
keeps 79.5\% of the left/right pairs and 64.0\% of the re-attachments, but only 4.60\% of the shifts to other body
locations. Applied to the estimate before the correction, the same rule keeps only 50.0\% of the recovered shifts and
abstains on 77.4\% of the others.
ATC also separates the shifts that a transform of the axes explains, the left/right pairs and re-attachments, from
the shifts to other body locations with an AUROC of 0.874. When the injected shifts of all 133 held-out
subjects are added to the real shifts, it separates the recovered shifts with an AUROC of 0.927.

\subsection{Ablation Study}
\label{sec:exp:ablation}

\begin{table}[t]
\caption{Ablation of the re-attachment correction on REALDISP (held-out subjects) and on the eight datasets (mean) with simulated re-attachment or none. Without rotation selection, the correction is normalization re-estimation alone (AdaBN), which never turns a unit and so leads when no unit was re-attached. Best in bold, second best underlined.}
\label{tab:ablation}
\scriptsize
\setlength{\tabcolsep}{0pt}
\begin{tabular*}{\linewidth}{@{\extracolsep{\fill}}lccc@{\hspace{2em}}c}
\toprule
& Real re-attachment & \multicolumn{2}{c}{Simulated re-attachment} & \\
\cmidrule(lr){2-2}\cmidrule(lr){3-4}
Variant & REALDISP & Axis-aligned & Arbitrary & No re-attachment \\
\midrule
w/o rotation selection & 0.808 & 0.374 & 0.370 & \textbf{0.883} \\
w/o normalization re-estimation & \underline{0.931} & 0.791 & 0.466 & 0.837 \\
w/o rotation restriction & 0.862 & \underline{0.824} & \underline{0.576} & 0.852 \\
\midrule
REFIT (ours) & \textbf{0.955} & \textbf{0.861} & \textbf{0.640} & \underline{0.876} \\
\bottomrule
\end{tabular*}
\end{table}

Table~\ref{tab:ablation} removes one component of the re-attachment correction at a time. The full correction is
the best of the variants in the table on real re-attachment and under both kinds of simulated re-attachment. When no unit was
re-attached, it stays 0.00722 below normalization re-estimation alone, because the selection occasionally turns a unit
that was not moved (Appendix~\ref{app:limits}).

\textbf{Without rotation selection}, normalization re-estimation alone cannot restore the turned axes. The statistics of each
channel are matched again, but gravity and the motion still arrive on the wrong axes. The accuracy under
axis-aligned re-attachment falls from 0.861 to 0.374.

\textbf{Without normalization re-estimation}, the selected rotation is applied but the model keeps the statistics of the training
subjects and of an exactly aligned strap. This costs most under arbitrary re-attachment (0.466 against 0.640), where
the closest of the 24 rotations leaves the largest residual rotation. The variant also lowers the accuracy on REALDISP (0.931
against 0.955). Replacing normalization re-estimation by feature alignment is on par with the full correction on REALDISP
(0.957) but lowers the mean under arbitrary re-attachment to 0.531.

\textbf{Without the restriction to rotations}, the search also admits reflections, some of which match the moments of a rotated unit
almost as well. On DSADS, HuGaDB and OPPORTUNITY, it picks a reflection for 21.7\% of the re-attached units. The accuracy
then drops on REALDISP from 0.955 to 0.862 and under axis-aligned re-attachment from 0.861 to 0.824.

The family also has to match the shift when the shift is a mirror. \textbf{Without reflections}, the search over the 24
rotations cannot express the mirror after a side switch. Selection then reaches 0.484 on average over the five datasets of
Table~\ref{tab:mirror}, against 0.705 over the 48 signed permutations.
Flipping the same axis on the accelerometer and the gyroscope treats the gyroscope as if it measured an ordinary vector.
In an additional run, this flip stays below the source model on four of the five datasets.

Identification and verification can be ablated in the same way. \textbf{Without identification}, always correcting a rotation
lowers the accuracy over all test cases (Section~\ref{sec:exp:identify}) from 0.794 to 0.607. \textbf{Without the correction before
verification}, the AUROC of ATC falls from 0.925 to 0.686 (Table~\ref{tab:verify}).

These results show that the family of transforms must match the shift. Rotation selection and normalization
re-estimation act together, and each is necessary once a unit has been re-attached. Correcting the axes before
verification makes the accuracy estimate far more informative.

\subsection{Data Efficiency and Cost}
\label{sec:exp:cost}

\head{Amount of unlabeled data.}
REFIT needs only a sample of the held-out subject's windows. We compute the rotation selection and the normalization statistics
from 64 windows drawn at random from the recording and apply the result to all windows. The accuracy under
axis-aligned re-attachment then comes within 0.0177 of using every window, and with 16 windows within 0.0738
(Appendix~\ref{app:data}). Sixty-four windows are about five minutes of signal on DSADS, whose windows last
five seconds, and about one minute on HuGaDB and OPPORTUNITY, whose windows last about one second. On HuGaDB, 16 random
windows, about 16 seconds of signal, already reach 0.931, against 0.960 with every window.
What matters is that the windows cover several activities. Windows taken from one continuous stretch cover few
activities, and REFIT then needs many more of them. On DSADS, a continuous stretch of 256 windows, about 21 minutes,
reaches only 0.501, against 0.851 for 256 windows drawn at random.

\head{Computational cost.}
The rotation selection reads only moments of the raw signal and takes less than two tenths of a second for 11,110
windows on one CPU thread. The complete calibration costs about as much as AdaBN, at most 1.15 times its time and at
most 3.02 seconds. SHOT, which adapts the feature extractor, needs 25.7 to 87.6 seconds
(Appendix~\ref{app:cost}).

\section{Discussion}
\label{sec:discussion}

The results above suggest how REFIT can run on a wearable device. We discuss when to run it and what it costs, how it
abstains and informs the user, how it handles several sensors, and which families of transforms suit a given device.

\head{When to run REFIT and what it costs.}
A placement shift arises when a device is put on again, so REFIT runs on the unlabeled windows recorded after that
moment. It needs only a sample of them, but the sample must cover several activities, because the windows of one
continuous stretch need many more (Section~\ref{sec:exp:cost}). A device can therefore collect windows in the
background, sample them over the activities of the day, and run REFIT once enough varied windows are available. For
most users whose sensor did not move, the selection keeps the identity and REFIT reduces to normalization
re-estimation. On average over such users, it stayed within 0.01 of AdaBN on the two datasets not used to design REFIT
(Section~\ref{sec:exp:reattach}). Besides the frozen model, REFIT stores for each unit the mean
acceleration and the two covariances of the training users, which the correction needs. Identification adds the mean
and spread of the statistics vector $m$ and the percentile below which the input is left unchanged, and verification adds one
ATC threshold per model. These are about a hundred numbers per unit. At run time, REFIT computes moments of the user's
windows, scores a few dozen candidate transforms with small matrix products, re-estimates the normalization statistics
in one pass and classifies the windows in one forward pass. It takes no gradient step, so it needs no memory for
training. The same frozen model serves every user, and the state kept per user is one transform per unit and the
re-estimated normalization statistics.

\head{Abstaining and informing the user.}
The accuracy below which REFIT abstains depends on the application. A tracker that counts active minutes tolerates more
errors than a monitor whose alerts reach a caregiver. ATC estimates the accuracy in the same units as such a
requirement, so the threshold can be stated as an accuracy. With the rule of Section~\ref{sec:exp:verify}, REFIT keeps
the correction for most of the recovered placement shifts and abstains mostly on shifts to other body locations, the case that no
transform of the axes can explain. When REFIT abstains, the device can ask the user to put the sensor back as it was
worn in training and run REFIT again on new windows. Because REFIT names the family of the shift, the device can also
tell the user which kind of correction it applied. For example, a mirror suggests that the watch now sits on the other
wrist, and a rotation suggests that a strap sensor was put back turned. A method that only adapts statistics or weights has no such description to offer. The user thus stays
informed about a correction that would otherwise happen out of sight.

\head{Several sensors.}
REFIT selects the transform of each unit from that unit's moments alone and then re-estimates the normalization
statistics once for all units. In a system with several sensors, it can therefore turn a strap that was put back
differently, keep the others, and report which of its sensors it turned. On REALDISP, it turns at least one
unit for 13 of the 15 subjects and keeps the others.

\head{Choosing the families.}
The families are a design choice for each device. Where only side switches are expected, the search over the 48
signed permutations covers every axis convention (Table~\ref{tab:mirror}), and the fixed mirror of Eq.~\eqref{eq:mirror}
applies directly once the convention of a device is known. A box-shaped sensor on a strap
calls for the 24 rotations, and the restriction to rotations matters, since admitting reflections lowers the accuracy
after a re-attachment (Table~\ref{tab:ablation}). When the kind of shift is unknown, identification over all six
families chooses the correction (Tables~\ref{tab:naming} and~\ref{tab:identify}). For accelerometer streams without gravity,
Appendix~\ref{app:gravity} evaluates a variant of the rotation selection.

\section{Conclusion}
\label{sec:conclusion}

In this work, we presented REFIT, a label-free input calibration that lets a frozen wearable model keep working when its
sensors are worn differently than in training. Unlike most test-time adaptation methods, which adjust statistics, features or
prototypes, REFIT treats the common placement shifts as transforms of the sensor axes and undoes them in front of the
model, whose weights stay fixed. It recognizes the kind of shift by fitting families of transforms to raw-signal moments. It fixes the
shift by applying the fitted transform and re-estimating the normalization statistics of the model. A label-free accuracy
estimate then tests the fixed model and decides whether to ask the user to re-wear the sensor. Across real
left/right sensor pairs and real and simulated re-attachment, REFIT outperforms label-free adaptation methods on every
dataset and restores most of the accuracy lost to re-attachment, at about the cost of re-estimating normalization
statistics. It names injected shifts far more reliably than the confidence selector.
After a correction over all signed permutations of the axes, a label-free accuracy estimate separates the recovered
placement shifts from the others. Extending the families to continuous orientations is a
natural next step. We believe this work provides a practical step toward wearable models that remain accurate however
their sensors are worn.

\section*{Generative AI Disclosure}
We used generative AI tools as assistants while writing analysis code and while editing the text of this paper. The
author reviewed and verified all code, results and text produced with their help, and takes full responsibility for
the content of the paper.

\bibliographystyle{ACM-Reference-Format}
\bibliography{references}


\begin{thebibliography}{97}


\ifx \showCODEN    \undefined \def \showCODEN     #1{\unskip}     \fi
\ifx \showISBNx    \undefined \def \showISBNx     #1{\unskip}     \fi
\ifx \showISBNxiii \undefined \def \showISBNxiii  #1{\unskip}     \fi
\ifx \showISSN     \undefined \def \showISSN      #1{\unskip}     \fi
\ifx \showLCCN     \undefined \def \showLCCN      #1{\unskip}     \fi
\ifx \shownote     \undefined \def \shownote      #1{#1}          \fi
\ifx \showarticletitle \undefined \def \showarticletitle #1{#1}   \fi
\ifx \showURL      \undefined \def \showURL       {\relax}        \fi
\providecommand\bibfield[2]{#2}
\providecommand\bibinfo[2]{#2}
\providecommand\natexlab[1]{#1}
\providecommand\showeprint[2][]{arXiv:#2}

\bibitem[Altun et~al\mbox{.}(2010)]%
        {dsads}
\bibfield{author}{\bibinfo{person}{Kerem Altun}, \bibinfo{person}{Billur
  Barshan}, {and} \bibinfo{person}{Orkun Tun{\c{c}}el}.}
  \bibinfo{year}{2010}\natexlab{}.
\newblock \showarticletitle{Comparative Study on Classifying Human Activities
  with Miniature Inertial and Magnetic Sensors}.
\newblock \bibinfo{journal}{\emph{Pattern Recognition}} \bibinfo{volume}{43},
  \bibinfo{number}{10} (\bibinfo{year}{2010}), \bibinfo{pages}{3605--3620}.
\newblock
\href{https://doi.org/10.1016/j.patcog.2010.04.019}{doi:\nolinkurl{10.1016/j.patcog.2010.04.019}}


\bibitem[Anguita et~al\mbox{.}(2013)]%
        {ucihar}
\bibfield{author}{\bibinfo{person}{Davide Anguita}, \bibinfo{person}{Alessandro
  Ghio}, \bibinfo{person}{Luca Oneto}, \bibinfo{person}{Xavier Parra}, {and}
  \bibinfo{person}{Jorge~L. Reyes-Ortiz}.} \bibinfo{year}{2013}\natexlab{}.
\newblock \showarticletitle{A Public Domain Dataset for Human Activity
  Recognition Using Smartphones}. In \bibinfo{booktitle}{\emph{21st European
  Symposium on Artificial Neural Networks, Computational Intelligence and
  Machine Learning (ESANN)}}. \bibinfo{publisher}{i6doc.com},
  \bibinfo{address}{Bruges, Belgium}, \bibinfo{pages}{437--442}.
\newblock
\showISBNx{978-2-87419-081-0}


\bibitem[Baek et~al\mbox{.}(2022)]%
        {baek2022agreement}
\bibfield{author}{\bibinfo{person}{Christina Baek}, \bibinfo{person}{Yiding
  Jiang}, \bibinfo{person}{Aditi Raghunathan}, {and} \bibinfo{person}{J.~Zico
  Kolter}.} \bibinfo{year}{2022}\natexlab{}.
\newblock \showarticletitle{Agreement-on-the-line: Predicting the Performance
  of Neural Networks under Distribution Shift}. In
  \bibinfo{booktitle}{\emph{Advances in Neural Information Processing Systems
  (NeurIPS)}}, Vol.~\bibinfo{volume}{35}. \bibinfo{publisher}{Curran
  Associates, Inc.}, \bibinfo{pages}{19274--19289}.
\newblock
\href{https://doi.org/10.52202/068431-1401}{doi:\nolinkurl{10.52202/068431-1401}}


\bibitem[Baek et~al\mbox{.}(2026)]%
        {baek2026trihar}
\bibfield{author}{\bibinfo{person}{Seungyeol Baek}, \bibinfo{person}{Yoonbyung
  Chai}, \bibinfo{person}{Yonghyeon Lee}, \bibinfo{person}{Sungjoon Choi},
  {and} \bibinfo{person}{Sungho Suh}.} \bibinfo{year}{2026}\natexlab{}.
\newblock \showarticletitle{Rotation-Invariant Multi-{IMU} Activity Recognition
  under Independent Per-Location Orientation Shifts}. In
  \bibinfo{booktitle}{\emph{Proceedings of the 2026 ACM International Symposium
  on Wearable Computers (ISWC)}}. \bibinfo{publisher}{ACM},
  \bibinfo{pages}{1--7}.
\newblock
\href{https://doi.org/10.1145/3830727.3834824}{doi:\nolinkurl{10.1145/3830727.3834824}}


\bibitem[Banos et~al\mbox{.}(2012)]%
        {realdisp}
\bibfield{author}{\bibinfo{person}{Oresti Banos}, \bibinfo{person}{Miguel
  Damas}, \bibinfo{person}{Hector Pomares}, \bibinfo{person}{Ignacio Rojas},
  \bibinfo{person}{Mate~Attila Toth}, {and} \bibinfo{person}{Oliver Amft}.}
  \bibinfo{year}{2012}\natexlab{}.
\newblock \showarticletitle{A Benchmark Dataset to Evaluate Sensor Displacement
  in Activity Recognition}. In \bibinfo{booktitle}{\emph{Proceedings of the
  2012 ACM Conference on Ubiquitous Computing (UbiComp '12)}}.
  \bibinfo{publisher}{ACM}, \bibinfo{pages}{1026--1035}.
\newblock
\href{https://doi.org/10.1145/2370216.2370437}{doi:\nolinkurl{10.1145/2370216.2370437}}


\bibitem[Banos et~al\mbox{.}(2014a)]%
        {mhealth}
\bibfield{author}{\bibinfo{person}{Oresti Banos}, \bibinfo{person}{Rafael
  Garcia}, \bibinfo{person}{Juan~A. Holgado-Terriza}, \bibinfo{person}{Miguel
  Damas}, \bibinfo{person}{Hector Pomares}, \bibinfo{person}{Ignacio Rojas},
  \bibinfo{person}{Alejandro Saez}, {and} \bibinfo{person}{Claudia
  Villalonga}.} \bibinfo{year}{2014}\natexlab{a}.
\newblock \showarticletitle{{mHealthDroid}: A Novel Framework for Agile
  Development of Mobile Health Applications}. In
  \bibinfo{booktitle}{\emph{Ambient Assisted Living and Daily Activities (IWAAL
  2014)}} \emph{(\bibinfo{series}{Lecture Notes in Computer Science},
  Vol.~\bibinfo{volume}{8868})}. \bibinfo{publisher}{Springer},
  \bibinfo{address}{Cham}, \bibinfo{pages}{91--98}.
\newblock
\href{https://doi.org/10.1007/978-3-319-13105-4_14}{doi:\nolinkurl{10.1007/978-3-319-13105-4_14}}


\bibitem[Banos et~al\mbox{.}(2014b)]%
        {banos2014displacement}
\bibfield{author}{\bibinfo{person}{Oresti Banos}, \bibinfo{person}{Mate~Attila
  Toth}, \bibinfo{person}{Miguel Damas}, \bibinfo{person}{Hector Pomares},
  {and} \bibinfo{person}{Ignacio Rojas}.} \bibinfo{year}{2014}\natexlab{b}.
\newblock \showarticletitle{Dealing with the Effects of Sensor Displacement in
  Wearable Activity Recognition}.
\newblock \bibinfo{journal}{\emph{Sensors}} \bibinfo{volume}{14},
  \bibinfo{number}{6} (\bibinfo{year}{2014}), \bibinfo{pages}{9995--10023}.
\newblock
\href{https://doi.org/10.3390/s140609995}{doi:\nolinkurl{10.3390/s140609995}}


\bibitem[Boudiaf et~al\mbox{.}(2022)]%
        {lame}
\bibfield{author}{\bibinfo{person}{Malik Boudiaf}, \bibinfo{person}{Romain
  Mueller}, \bibinfo{person}{Ismail {Ben Ayed}}, {and} \bibinfo{person}{Luca
  Bertinetto}.} \bibinfo{year}{2022}\natexlab{}.
\newblock \showarticletitle{Parameter-free Online Test-time Adaptation}. In
  \bibinfo{booktitle}{\emph{Proceedings of the IEEE/CVF Conference on Computer
  Vision and Pattern Recognition (CVPR)}}. \bibinfo{publisher}{IEEE},
  \bibinfo{pages}{8334--8343}.
\newblock
\href{https://doi.org/10.1109/CVPR52688.2022.00816}{doi:\nolinkurl{10.1109/CVPR52688.2022.00816}}


\bibitem[Buchan et~al\mbox{.}(2020)]%
        {buchan2020comparison}
\bibfield{author}{\bibinfo{person}{Duncan~S. Buchan}, \bibinfo{person}{Lynne~M.
  Boddy}, {and} \bibinfo{person}{Gillian McLellan}.}
  \bibinfo{year}{2020}\natexlab{}.
\newblock \showarticletitle{Comparison of Free-Living and Laboratory Activity
  Outcomes from {ActiGraph} Accelerometers Worn on the Dominant and
  Non-Dominant Wrists}.
\newblock \bibinfo{journal}{\emph{Measurement in Physical Education and
  Exercise Science}} \bibinfo{volume}{24}, \bibinfo{number}{4}
  (\bibinfo{year}{2020}), \bibinfo{pages}{247--257}.
\newblock
\href{https://doi.org/10.1080/1091367X.2020.1801441}{doi:\nolinkurl{10.1080/1091367X.2020.1801441}}


\bibitem[Chang et~al\mbox{.}(2020)]%
        {chang2020systematic}
\bibfield{author}{\bibinfo{person}{Youngjae Chang}, \bibinfo{person}{Akhil
  Mathur}, \bibinfo{person}{Anton Isopoussu}, \bibinfo{person}{Junehwa Song},
  {and} \bibinfo{person}{Fahim Kawsar}.} \bibinfo{year}{2020}\natexlab{}.
\newblock \showarticletitle{A Systematic Study of Unsupervised Domain
  Adaptation for Robust Human-Activity Recognition}.
\newblock \bibinfo{journal}{\emph{Proceedings of the ACM on Interactive,
  Mobile, Wearable and Ubiquitous Technologies}} \bibinfo{volume}{4},
  \bibinfo{number}{1} (\bibinfo{year}{2020}), \bibinfo{pages}{1--30}.
\newblock
\href{https://doi.org/10.1145/3380985}{doi:\nolinkurl{10.1145/3380985}}


\bibitem[Chavarriaga et~al\mbox{.}(2013a)]%
        {chavarriaga2013unsupervised}
\bibfield{author}{\bibinfo{person}{Ricardo Chavarriaga},
  \bibinfo{person}{Hamidreza Bayati}, {and} \bibinfo{person}{Jos{\'e} del~R.
  Mill{\'a}n}.} \bibinfo{year}{2013}\natexlab{a}.
\newblock \showarticletitle{Unsupervised Adaptation for Acceleration-Based
  Activity Recognition: Robustness to Sensor Displacement and Rotation}.
\newblock \bibinfo{journal}{\emph{Personal and Ubiquitous Computing}}
  \bibinfo{volume}{17}, \bibinfo{number}{3} (\bibinfo{year}{2013}),
  \bibinfo{pages}{479--490}.
\newblock
\href{https://doi.org/10.1007/s00779-011-0493-y}{doi:\nolinkurl{10.1007/s00779-011-0493-y}}


\bibitem[Chavarriaga et~al\mbox{.}(2013b)]%
        {opportunity}
\bibfield{author}{\bibinfo{person}{Ricardo Chavarriaga}, \bibinfo{person}{Hesam
  Sagha}, \bibinfo{person}{Alberto Calatroni}, \bibinfo{person}{Sundara~Tejaswi
  Digumarti}, \bibinfo{person}{Gerhard Tr{\"o}ster}, \bibinfo{person}{Jos{\'e}
  del~R. Mill{\'a}n}, {and} \bibinfo{person}{Daniel Roggen}.}
  \bibinfo{year}{2013}\natexlab{b}.
\newblock \showarticletitle{The {Opportunity} Challenge: A Benchmark Database
  for On-Body Sensor-Based Activity Recognition}.
\newblock \bibinfo{journal}{\emph{Pattern Recognition Letters}}
  \bibinfo{volume}{34}, \bibinfo{number}{15} (\bibinfo{year}{2013}),
  \bibinfo{pages}{2033--2042}.
\newblock
\href{https://doi.org/10.1016/j.patrec.2012.12.014}{doi:\nolinkurl{10.1016/j.patrec.2012.12.014}}


\bibitem[Chen et~al\mbox{.}(2021a)]%
        {chen2021detecting}
\bibfield{author}{\bibinfo{person}{Jiefeng Chen}, \bibinfo{person}{Frederick
  Liu}, \bibinfo{person}{Besim Avci}, \bibinfo{person}{Xi Wu},
  \bibinfo{person}{Yingyu Liang}, {and} \bibinfo{person}{Somesh Jha}.}
  \bibinfo{year}{2021}\natexlab{a}.
\newblock \showarticletitle{Detecting Errors and Estimating Accuracy on
  Unlabeled Data with Self-training Ensembles}. In
  \bibinfo{booktitle}{\emph{Advances in Neural Information Processing Systems
  (NeurIPS)}}, Vol.~\bibinfo{volume}{34}. \bibinfo{publisher}{Curran
  Associates, Inc.}, \bibinfo{pages}{14980--14992}.
\newblock
\urldef\tempurl%
\url{https://proceedings.neurips.cc/paper/2021/hash/7dd3ed2e12d7967b656d156d50308263-Abstract.html}
\showURL{%
\tempurl}


\bibitem[Chen et~al\mbox{.}(2021b)]%
        {chen2021survey}
\bibfield{author}{\bibinfo{person}{Kaixuan Chen}, \bibinfo{person}{Dalin
  Zhang}, \bibinfo{person}{Lina Yao}, \bibinfo{person}{Bin Guo},
  \bibinfo{person}{Zhiwen Yu}, {and} \bibinfo{person}{Yunhao Liu}.}
  \bibinfo{year}{2021}\natexlab{b}.
\newblock \showarticletitle{Deep Learning for Sensor-based Human Activity
  Recognition: Overview, Challenges, and Opportunities}.
\newblock \bibinfo{journal}{\emph{Comput. Surveys}} \bibinfo{volume}{54},
  \bibinfo{number}{4}, Article \bibinfo{articleno}{77} (\bibinfo{year}{2021}),
  \bibinfo{numpages}{40}~pages.
\newblock
\href{https://doi.org/10.1145/3447744}{doi:\nolinkurl{10.1145/3447744}}


\bibitem[Chen et~al\mbox{.}(2019)]%
        {wang2018crossposition}
\bibfield{author}{\bibinfo{person}{Yiqiang Chen}, \bibinfo{person}{Jindong
  Wang}, \bibinfo{person}{Meiyu Huang}, {and} \bibinfo{person}{Han Yu}.}
  \bibinfo{year}{2019}\natexlab{}.
\newblock \showarticletitle{Cross-position Activity Recognition with Stratified
  Transfer Learning}.
\newblock \bibinfo{journal}{\emph{Pervasive and Mobile Computing}}
  \bibinfo{volume}{57} (\bibinfo{year}{2019}), \bibinfo{pages}{1--13}.
\newblock
\href{https://doi.org/10.1016/j.pmcj.2019.04.004}{doi:\nolinkurl{10.1016/j.pmcj.2019.04.004}}


\bibitem[Chereshnev and Kert{\'e}sz-Farkas(2018)]%
        {hugadb}
\bibfield{author}{\bibinfo{person}{Roman Chereshnev} {and}
  \bibinfo{person}{Attila Kert{\'e}sz-Farkas}.}
  \bibinfo{year}{2018}\natexlab{}.
\newblock \showarticletitle{{HuGaDB}: Human Gait Database for Activity
  Recognition from Wearable Inertial Sensor Networks}. In
  \bibinfo{booktitle}{\emph{Analysis of Images, Social Networks and Texts (AIST
  2017)}} \emph{(\bibinfo{series}{Lecture Notes in Computer Science},
  Vol.~\bibinfo{volume}{10716})}. \bibinfo{publisher}{Springer},
  \bibinfo{pages}{131--141}.
\newblock
\href{https://doi.org/10.1007/978-3-319-73013-4_12}{doi:\nolinkurl{10.1007/978-3-319-73013-4_12}}


\bibitem[Cohen and Welling(2016)]%
        {cohen2016group}
\bibfield{author}{\bibinfo{person}{Taco~S. Cohen} {and} \bibinfo{person}{Max
  Welling}.} \bibinfo{year}{2016}\natexlab{}.
\newblock \showarticletitle{Group Equivariant Convolutional Networks}. In
  \bibinfo{booktitle}{\emph{Proceedings of the 33rd International Conference on
  Machine Learning (ICML)}} \emph{(\bibinfo{series}{Proceedings of Machine
  Learning Research}, Vol.~\bibinfo{volume}{48})}. \bibinfo{pages}{2990--2999}.
\newblock
\urldef\tempurl%
\url{http://proceedings.mlr.press/v48/cohenc16.html}
\showURL{%
\tempurl}


\bibitem[Deng and Zheng(2021)]%
        {deng2021labels}
\bibfield{author}{\bibinfo{person}{Weijian Deng} {and} \bibinfo{person}{Liang
  Zheng}.} \bibinfo{year}{2021}\natexlab{}.
\newblock \showarticletitle{Are Labels Always Necessary for Classifier Accuracy
  Evaluation?}. In \bibinfo{booktitle}{\emph{Proceedings of the IEEE/CVF
  Conference on Computer Vision and Pattern Recognition (CVPR)}}.
  \bibinfo{pages}{15069--15078}.
\newblock


\bibitem[F{\"o}rster et~al\mbox{.}(2009)]%
        {forster2009unsupervised}
\bibfield{author}{\bibinfo{person}{Kilian F{\"o}rster}, \bibinfo{person}{Daniel
  Roggen}, {and} \bibinfo{person}{Gerhard Tr{\"o}ster}.}
  \bibinfo{year}{2009}\natexlab{}.
\newblock \showarticletitle{Unsupervised Classifier Self-Calibration through
  Repeated Context Occurences: Is there Robustness against Sensor Displacement
  to Gain?}. In \bibinfo{booktitle}{\emph{2009 International Symposium on
  Wearable Computers (ISWC)}}. \bibinfo{publisher}{IEEE},
  \bibinfo{pages}{77--84}.
\newblock
\href{https://doi.org/10.1109/ISWC.2009.12}{doi:\nolinkurl{10.1109/ISWC.2009.12}}


\bibitem[Friedman et~al\mbox{.}(2012)]%
        {friedman2012orientation}
\bibfield{author}{\bibinfo{person}{Ascher Friedman},
  \bibinfo{person}{Nabil~Hajj Chehade}, \bibinfo{person}{Chieh Chien}, {and}
  \bibinfo{person}{Greg Pottie}.} \bibinfo{year}{2012}\natexlab{}.
\newblock \showarticletitle{Estimation of Accelerometer Orientation for
  Activity Recognition}. In \bibinfo{booktitle}{\emph{2012 Annual International
  Conference of the IEEE Engineering in Medicine and Biology Society (EMBC)}}.
  \bibinfo{pages}{2076--2079}.
\newblock
\href{https://doi.org/10.1109/EMBC.2012.6346368}{doi:\nolinkurl{10.1109/EMBC.2012.6346368}}


\bibitem[Gao et~al\mbox{.}(2023)]%
        {gao2023dda}
\bibfield{author}{\bibinfo{person}{Jin Gao}, \bibinfo{person}{Jialing Zhang},
  \bibinfo{person}{Xihui Liu}, \bibinfo{person}{Trevor Darrell},
  \bibinfo{person}{Evan Shelhamer}, {and} \bibinfo{person}{Dequan Wang}.}
  \bibinfo{year}{2023}\natexlab{}.
\newblock \showarticletitle{Back to the Source: Diffusion-Driven Adaptation to
  Test-Time Corruption}. In \bibinfo{booktitle}{\emph{Proceedings of the
  IEEE/CVF Conference on Computer Vision and Pattern Recognition (CVPR)}}.
  \bibinfo{publisher}{IEEE}, \bibinfo{pages}{11786--11796}.
\newblock
\href{https://doi.org/10.1109/CVPR52729.2023.01134}{doi:\nolinkurl{10.1109/CVPR52729.2023.01134}}


\bibitem[Garg et~al\mbox{.}(2022)]%
        {garg2022atc}
\bibfield{author}{\bibinfo{person}{Saurabh Garg}, \bibinfo{person}{Sivaraman
  Balakrishnan}, \bibinfo{person}{Zachary~C. Lipton}, \bibinfo{person}{Behnam
  Neyshabur}, {and} \bibinfo{person}{Hanie Sedghi}.}
  \bibinfo{year}{2022}\natexlab{}.
\newblock \showarticletitle{Leveraging Unlabeled Data to Predict
  Out-of-Distribution Performance}. In \bibinfo{booktitle}{\emph{International
  Conference on Learning Representations (ICLR)}}.
\newblock
\showeprint[arxiv]{2201.04234}


\bibitem[Geifman and El-Yaniv(2017)]%
        {geifman2017selective}
\bibfield{author}{\bibinfo{person}{Yonatan Geifman} {and} \bibinfo{person}{Ran
  El-Yaniv}.} \bibinfo{year}{2017}\natexlab{}.
\newblock \showarticletitle{Selective Classification for Deep Neural Networks}.
  In \bibinfo{booktitle}{\emph{Advances in Neural Information Processing
  Systems (NeurIPS)}}, Vol.~\bibinfo{volume}{30}. \bibinfo{publisher}{Curran
  Associates, Inc.}, \bibinfo{pages}{4878--4887}.
\newblock


\bibitem[Gjoreski et~al\mbox{.}(2016)]%
        {gjoreski2016wrist}
\bibfield{author}{\bibinfo{person}{Martin Gjoreski}, \bibinfo{person}{Hristijan
  Gjoreski}, \bibinfo{person}{Mitja Lu{\v{s}}trek}, {and}
  \bibinfo{person}{Matja{\v{z}} Gams}.} \bibinfo{year}{2016}\natexlab{}.
\newblock \showarticletitle{How Accurately Can Your Wrist Device Recognize
  Daily Activities and Detect Falls?}
\newblock \bibinfo{journal}{\emph{Sensors}} \bibinfo{volume}{16},
  \bibinfo{number}{6} (\bibinfo{year}{2016}), \bibinfo{pages}{800}.
\newblock
\href{https://doi.org/10.3390/s16060800}{doi:\nolinkurl{10.3390/s16060800}}


\bibitem[Gong et~al\mbox{.}(2022)]%
        {gong2022note}
\bibfield{author}{\bibinfo{person}{Taesik Gong}, \bibinfo{person}{Jongheon
  Jeong}, \bibinfo{person}{Taewon Kim}, \bibinfo{person}{Yewon Kim},
  \bibinfo{person}{Jinwoo Shin}, {and} \bibinfo{person}{Sung-Ju Lee}.}
  \bibinfo{year}{2022}\natexlab{}.
\newblock \showarticletitle{{NOTE}: Robust Continual Test-time Adaptation
  Against Temporal Correlation}. In \bibinfo{booktitle}{\emph{Advances in
  Neural Information Processing Systems (NeurIPS)}}, Vol.~\bibinfo{volume}{35}.
  \bibinfo{publisher}{Curran Associates, Inc.}, \bibinfo{pages}{27253--27266}.
\newblock
\href{https://doi.org/10.52202/068431-1976}{doi:\nolinkurl{10.52202/068431-1976}}


\bibitem[Gong et~al\mbox{.}(2023)]%
        {gong2023dapper}
\bibfield{author}{\bibinfo{person}{Taesik Gong}, \bibinfo{person}{Yewon Kim},
  \bibinfo{person}{Adiba Orzikulova}, \bibinfo{person}{Yunxin Liu},
  \bibinfo{person}{Sung~Ju Hwang}, \bibinfo{person}{Jinwoo Shin}, {and}
  \bibinfo{person}{Sung-Ju Lee}.} \bibinfo{year}{2023}\natexlab{}.
\newblock \showarticletitle{{DAPPER}: Label-Free Performance Estimation after
  Personalization for Heterogeneous Mobile Sensing}.
\newblock \bibinfo{journal}{\emph{Proceedings of the ACM on Interactive,
  Mobile, Wearable and Ubiquitous Technologies}} \bibinfo{volume}{7},
  \bibinfo{number}{2} (\bibinfo{year}{2023}), \bibinfo{pages}{1--27}.
\newblock
\href{https://doi.org/10.1145/3596256}{doi:\nolinkurl{10.1145/3596256}}


\bibitem[Gong et~al\mbox{.}(2019)]%
        {gong2019metasense}
\bibfield{author}{\bibinfo{person}{Taesik Gong}, \bibinfo{person}{Yeonsu Kim},
  \bibinfo{person}{Jinwoo Shin}, {and} \bibinfo{person}{Sung-Ju Lee}.}
  \bibinfo{year}{2019}\natexlab{}.
\newblock \showarticletitle{{MetaSense}: Few-Shot Adaptation to Untrained
  Conditions in Deep Mobile Sensing}. In \bibinfo{booktitle}{\emph{Proceedings
  of the 17th Conference on Embedded Networked Sensor Systems (SenSys)}}.
  \bibinfo{publisher}{ACM}, \bibinfo{pages}{110--123}.
\newblock
\href{https://doi.org/10.1145/3356250.3360020}{doi:\nolinkurl{10.1145/3356250.3360020}}


\bibitem[Guillory et~al\mbox{.}(2021)]%
        {guillory2021doc}
\bibfield{author}{\bibinfo{person}{Devin Guillory}, \bibinfo{person}{Vaishaal
  Shankar}, \bibinfo{person}{Sayna Ebrahimi}, \bibinfo{person}{Trevor Darrell},
  {and} \bibinfo{person}{Ludwig Schmidt}.} \bibinfo{year}{2021}\natexlab{}.
\newblock \showarticletitle{Predicting with Confidence on Unseen
  Distributions}. In \bibinfo{booktitle}{\emph{Proceedings of the IEEE/CVF
  International Conference on Computer Vision (ICCV)}}.
  \bibinfo{pages}{1134--1144}.
\newblock


\bibitem[Henpraserttae et~al\mbox{.}(2011)]%
        {henpraserttae2011orientation}
\bibfield{author}{\bibinfo{person}{Apiwat Henpraserttae},
  \bibinfo{person}{Surapa Thiemjarus}, {and} \bibinfo{person}{Sanparith
  Marukatat}.} \bibinfo{year}{2011}\natexlab{}.
\newblock \showarticletitle{Accurate Activity Recognition Using a Mobile Phone
  Regardless of Device Orientation and Location}. In
  \bibinfo{booktitle}{\emph{2011 International Conference on Body Sensor
  Networks (BSN)}}. \bibinfo{pages}{41--46}.
\newblock
\href{https://doi.org/10.1109/BSN.2011.8}{doi:\nolinkurl{10.1109/BSN.2011.8}}


\bibitem[Heydarian et~al\mbox{.}(2020)]%
        {heydarian2020intake}
\bibfield{author}{\bibinfo{person}{Hamid Heydarian},
  \bibinfo{person}{Philipp~V. Rouast}, \bibinfo{person}{Marc T.~P. Adam},
  \bibinfo{person}{Tracy Burrows}, \bibinfo{person}{Clare~E. Collins}, {and}
  \bibinfo{person}{Megan~E. Rollo}.} \bibinfo{year}{2020}\natexlab{}.
\newblock \showarticletitle{Deep Learning for Intake Gesture Detection From
  Wrist-Worn Inertial Sensors: The Effects of Data Preprocessing, Sensor
  Modalities, and Sensor Positions}.
\newblock \bibinfo{journal}{\emph{IEEE Access}}  \bibinfo{volume}{8}
  (\bibinfo{year}{2020}), \bibinfo{pages}{164936--164949}.
\newblock
\href{https://doi.org/10.1109/ACCESS.2020.3022042}{doi:\nolinkurl{10.1109/ACCESS.2020.3022042}}


\bibitem[Holm(1979)]%
        {holm1979}
\bibfield{author}{\bibinfo{person}{Sture Holm}.}
  \bibinfo{year}{1979}\natexlab{}.
\newblock \showarticletitle{A Simple Sequentially Rejective Multiple Test
  Procedure}.
\newblock \bibinfo{journal}{\emph{Scandinavian Journal of Statistics}}
  \bibinfo{volume}{6}, \bibinfo{number}{2} (\bibinfo{year}{1979}),
  \bibinfo{pages}{65--70}.
\newblock


\bibitem[Hong et~al\mbox{.}(2024)]%
        {hong2024crosshar}
\bibfield{author}{\bibinfo{person}{Zhiqing Hong}, \bibinfo{person}{Zelong Li},
  \bibinfo{person}{Shuxin Zhong}, \bibinfo{person}{Wenjun Lyu},
  \bibinfo{person}{Haotian Wang}, \bibinfo{person}{Yi Ding},
  \bibinfo{person}{Tian He}, {and} \bibinfo{person}{Desheng Zhang}.}
  \bibinfo{year}{2024}\natexlab{}.
\newblock \showarticletitle{{CrossHAR}: Generalizing Cross-dataset Human
  Activity Recognition via Hierarchical Self-Supervised Pretraining}.
\newblock \bibinfo{journal}{\emph{Proceedings of the ACM on Interactive,
  Mobile, Wearable and Ubiquitous Technologies}} \bibinfo{volume}{8},
  \bibinfo{number}{2} (\bibinfo{year}{2024}), \bibinfo{pages}{1--26}.
\newblock
\href{https://doi.org/10.1145/3659597}{doi:\nolinkurl{10.1145/3659597}}


\bibitem[Ioffe and Szegedy(2015)]%
        {ioffe2015batchnorm}
\bibfield{author}{\bibinfo{person}{Sergey Ioffe} {and}
  \bibinfo{person}{Christian Szegedy}.} \bibinfo{year}{2015}\natexlab{}.
\newblock \showarticletitle{Batch Normalization: Accelerating Deep Network
  Training by Reducing Internal Covariate Shift}. In
  \bibinfo{booktitle}{\emph{Proceedings of the 32nd International Conference on
  Machine Learning (ICML)}} \emph{(\bibinfo{series}{Proceedings of Machine
  Learning Research}, Vol.~\bibinfo{volume}{37})}. \bibinfo{publisher}{PMLR},
  \bibinfo{pages}{448--456}.
\newblock


\bibitem[Iwasawa and Matsuo(2021)]%
        {t3a}
\bibfield{author}{\bibinfo{person}{Yusuke Iwasawa} {and}
  \bibinfo{person}{Yutaka Matsuo}.} \bibinfo{year}{2021}\natexlab{}.
\newblock \showarticletitle{Test-Time Classifier Adjustment Module for
  Model-Agnostic Domain Generalization}. In \bibinfo{booktitle}{\emph{Advances
  in Neural Information Processing Systems (NeurIPS)}},
  Vol.~\bibinfo{volume}{34}. \bibinfo{pages}{2427--2440}.
\newblock


\bibitem[Jayanth et~al\mbox{.}(2025)]%
        {eqnio}
\bibfield{author}{\bibinfo{person}{Royina~Karegoudra Jayanth},
  \bibinfo{person}{Yinshuang Xu}, \bibinfo{person}{Ziyun Wang},
  \bibinfo{person}{Evangelos Chatzipantazis}, \bibinfo{person}{Kostas
  Daniilidis}, {and} \bibinfo{person}{Daniel Gehrig}.}
  \bibinfo{year}{2025}\natexlab{}.
\newblock \showarticletitle{{EqNIO}: Subequivariant Neural Inertial Odometry}.
  In \bibinfo{booktitle}{\emph{International Conference on Learning
  Representations (ICLR)}}.
\newblock


\bibitem[Kaba et~al\mbox{.}(2023)]%
        {kaba2023canonicalization}
\bibfield{author}{\bibinfo{person}{S{\'e}kou-Oumar Kaba},
  \bibinfo{person}{Arnab~Kumar Mondal}, \bibinfo{person}{Yan Zhang},
  \bibinfo{person}{Yoshua Bengio}, {and} \bibinfo{person}{Siamak Ravanbakhsh}.}
  \bibinfo{year}{2023}\natexlab{}.
\newblock \showarticletitle{Equivariance with Learned Canonicalization
  Functions}. In \bibinfo{booktitle}{\emph{Proceedings of the 40th
  International Conference on Machine Learning (ICML)}}
  \emph{(\bibinfo{series}{Proceedings of Machine Learning Research},
  Vol.~\bibinfo{volume}{202})}. \bibinfo{pages}{15546--15566}.
\newblock


\bibitem[Kang et~al\mbox{.}(2023)]%
        {kang2023sfadapter}
\bibfield{author}{\bibinfo{person}{Hua Kang}, \bibinfo{person}{Qingyong Hu},
  {and} \bibinfo{person}{Qian Zhang}.} \bibinfo{year}{2023}\natexlab{}.
\newblock \showarticletitle{{SF-Adapter}: Computational-Efficient Source-Free
  Domain Adaptation for Human Activity Recognition}.
\newblock \bibinfo{journal}{\emph{Proceedings of the ACM on Interactive,
  Mobile, Wearable and Ubiquitous Technologies}} \bibinfo{volume}{7},
  \bibinfo{number}{4} (\bibinfo{year}{2023}), \bibinfo{pages}{1--23}.
\newblock
\href{https://doi.org/10.1145/3631428}{doi:\nolinkurl{10.1145/3631428}}


\bibitem[Karagiannaki et~al\mbox{.}(2016)]%
        {forthtrace}
\bibfield{author}{\bibinfo{person}{Katerina Karagiannaki},
  \bibinfo{person}{Athanasia Panousopoulou}, {and} \bibinfo{person}{Panagiotis
  Tsakalides}.} \bibinfo{year}{2016}\natexlab{}.
\newblock \bibinfo{title}{The {FORTH-TRACE} Dataset for Human Activity
  Recognition of Simple Activities and Postural Transitions Using a Body Area
  Network}.
\newblock \bibinfo{howpublished}{Zenodo}.
\newblock
\href{https://doi.org/10.5281/zenodo.841301}{doi:\nolinkurl{10.5281/zenodo.841301}}


\bibitem[Khazem and Serouis(2026)]%
        {khazem2026caster}
\bibfield{author}{\bibinfo{person}{Salim Khazem} {and}
  \bibinfo{person}{Ibrahim~Mohamed Serouis}.} \bibinfo{year}{2026}\natexlab{}.
\newblock \bibinfo{title}{Adapting Without Gradients: Affine Statistics
  Transport and What Its Certificate Can Tell You}.
\newblock
\showeprint[arxiv]{2609.00374}


\bibitem[Kunze and Lukowicz(2008)]%
        {kunze2008displacement}
\bibfield{author}{\bibinfo{person}{Kai Kunze} {and} \bibinfo{person}{Paul
  Lukowicz}.} \bibinfo{year}{2008}\natexlab{}.
\newblock \showarticletitle{Dealing with Sensor Displacement in Motion-Based
  Onbody Activity Recognition Systems}. In
  \bibinfo{booktitle}{\emph{Proceedings of the 10th International Conference on
  Ubiquitous Computing (UbiComp '08)}}. \bibinfo{publisher}{ACM},
  \bibinfo{pages}{20--29}.
\newblock
\href{https://doi.org/10.1145/1409635.1409639}{doi:\nolinkurl{10.1145/1409635.1409639}}


\bibitem[Kunze and Lukowicz(2014)]%
        {kunze2014placement}
\bibfield{author}{\bibinfo{person}{Kai Kunze} {and} \bibinfo{person}{Paul
  Lukowicz}.} \bibinfo{year}{2014}\natexlab{}.
\newblock \showarticletitle{Sensor Placement Variations in Wearable Activity
  Recognition}.
\newblock \bibinfo{journal}{\emph{IEEE Pervasive Computing}}
  \bibinfo{volume}{13}, \bibinfo{number}{4} (\bibinfo{year}{2014}),
  \bibinfo{pages}{32--41}.
\newblock
\href{https://doi.org/10.1109/MPRV.2014.73}{doi:\nolinkurl{10.1109/MPRV.2014.73}}


\bibitem[Kunze et~al\mbox{.}(2005)]%
        {kunze2005where}
\bibfield{author}{\bibinfo{person}{Kai Kunze}, \bibinfo{person}{Paul Lukowicz},
  \bibinfo{person}{Holger Junker}, {and} \bibinfo{person}{Gerhard
  Tr{\"o}ster}.} \bibinfo{year}{2005}\natexlab{}.
\newblock \showarticletitle{Where am {I}: Recognizing On-body Positions of
  Wearable Sensors}. In \bibinfo{booktitle}{\emph{Location- and
  Context-Awareness (LoCA 2005)}} \emph{(\bibinfo{series}{Lecture Notes in
  Computer Science}, Vol.~\bibinfo{volume}{3479})},
  \bibfield{editor}{\bibinfo{person}{Thomas Strang} {and}
  \bibinfo{person}{Claudia Linnhoff-Popien}} (Eds.).
  \bibinfo{publisher}{Springer}, \bibinfo{pages}{264--275}.
\newblock
\href{https://doi.org/10.1007/11426646_25}{doi:\nolinkurl{10.1007/11426646_25}}


\bibitem[Kunze et~al\mbox{.}(2009)]%
        {kunze2009which}
\bibfield{author}{\bibinfo{person}{Kai Kunze}, \bibinfo{person}{Paul Lukowicz},
  \bibinfo{person}{Kurt Partridge}, {and} \bibinfo{person}{Bo Begole}.}
  \bibinfo{year}{2009}\natexlab{}.
\newblock \showarticletitle{Which Way Am {I} Facing: Inferring Horizontal
  Device Orientation from an Accelerometer Signal}. In
  \bibinfo{booktitle}{\emph{2009 International Symposium on Wearable Computers
  (ISWC)}}. \bibinfo{publisher}{IEEE}, \bibinfo{pages}{149--150}.
\newblock
\href{https://doi.org/10.1109/ISWC.2009.33}{doi:\nolinkurl{10.1109/ISWC.2009.33}}


\bibitem[Kyritsis et~al\mbox{.}(2021)]%
        {kyritsis2021eating}
\bibfield{author}{\bibinfo{person}{Konstantinos Kyritsis},
  \bibinfo{person}{Christos Diou}, {and} \bibinfo{person}{Anastasios
  Delopoulos}.} \bibinfo{year}{2021}\natexlab{}.
\newblock \showarticletitle{A Data Driven End-to-End Approach for In-the-Wild
  Monitoring of Eating Behavior Using Smartwatches}.
\newblock \bibinfo{journal}{\emph{IEEE Journal of Biomedical and Health
  Informatics}} \bibinfo{volume}{25}, \bibinfo{number}{1}
  (\bibinfo{year}{2021}), \bibinfo{pages}{22--34}.
\newblock
\href{https://doi.org/10.1109/JBHI.2020.2984907}{doi:\nolinkurl{10.1109/JBHI.2020.2984907}}


\bibitem[Lee et~al\mbox{.}(2024)]%
        {lee2024aetta}
\bibfield{author}{\bibinfo{person}{Taeckyung Lee}, \bibinfo{person}{Sorn
  Chottananurak}, \bibinfo{person}{Taesik Gong}, {and} \bibinfo{person}{Sung-Ju
  Lee}.} \bibinfo{year}{2024}\natexlab{}.
\newblock \showarticletitle{{AETTA}: Label-Free Accuracy Estimation for
  Test-Time Adaptation}. In \bibinfo{booktitle}{\emph{Proceedings of the
  IEEE/CVF Conference on Computer Vision and Pattern Recognition (CVPR)}}.
  \bibinfo{pages}{28643--28652}.
\newblock
\href{https://doi.org/10.1109/CVPR52733.2024.02706}{doi:\nolinkurl{10.1109/CVPR52733.2024.02706}}


\bibitem[Li et~al\mbox{.}(2026)]%
        {li2026pitta}
\bibfield{author}{\bibinfo{person}{Changyu Li}, \bibinfo{person}{Lu Wang},
  \bibinfo{person}{Ming Lei}, \bibinfo{person}{Jiashen Liu},
  \bibinfo{person}{Yichen Zhang}, \bibinfo{person}{Kaishun Wu}, {and}
  \bibinfo{person}{Fei Luo}.} \bibinfo{year}{2026}\natexlab{}.
\newblock \bibinfo{title}{{PI-TTA}: Physics-Informed Source-Free Test-Time
  Adaptation for Robust Human Activity Recognition on Mobile Devices}.
\newblock
\showeprint[arxiv]{2604.25435}


\bibitem[Li et~al\mbox{.}(2018)]%
        {adabn}
\bibfield{author}{\bibinfo{person}{Yanghao Li}, \bibinfo{person}{Naiyan Wang},
  \bibinfo{person}{Jianping Shi}, \bibinfo{person}{Xiaodi Hou}, {and}
  \bibinfo{person}{Jiaying Liu}.} \bibinfo{year}{2018}\natexlab{}.
\newblock \showarticletitle{Adaptive Batch Normalization for Practical Domain
  Adaptation}.
\newblock \bibinfo{journal}{\emph{Pattern Recognition}}  \bibinfo{volume}{80}
  (\bibinfo{year}{2018}), \bibinfo{pages}{109--117}.
\newblock
\href{https://doi.org/10.1016/j.patcog.2018.03.005}{doi:\nolinkurl{10.1016/j.patcog.2018.03.005}}


\bibitem[Liang et~al\mbox{.}(2025)]%
        {liang2024ttasurvey}
\bibfield{author}{\bibinfo{person}{Jian Liang}, \bibinfo{person}{Ran He}, {and}
  \bibinfo{person}{Tieniu Tan}.} \bibinfo{year}{2025}\natexlab{}.
\newblock \showarticletitle{A Comprehensive Survey on Test-Time Adaptation
  under Distribution Shifts}.
\newblock \bibinfo{journal}{\emph{International Journal of Computer Vision}}
  \bibinfo{volume}{133}, \bibinfo{number}{1} (\bibinfo{year}{2025}),
  \bibinfo{pages}{31--64}.
\newblock
\href{https://doi.org/10.1007/s11263-024-02181-w}{doi:\nolinkurl{10.1007/s11263-024-02181-w}}


\bibitem[Liang et~al\mbox{.}(2020)]%
        {shot}
\bibfield{author}{\bibinfo{person}{Jian Liang}, \bibinfo{person}{Dapeng Hu},
  {and} \bibinfo{person}{Jiashi Feng}.} \bibinfo{year}{2020}\natexlab{}.
\newblock \showarticletitle{Do We Really Need to Access the Source Data?
  {S}ource Hypothesis Transfer for Unsupervised Domain Adaptation}. In
  \bibinfo{booktitle}{\emph{Proceedings of the 37th International Conference on
  Machine Learning (ICML)}} \emph{(\bibinfo{series}{Proceedings of Machine
  Learning Research}, Vol.~\bibinfo{volume}{119})}. \bibinfo{publisher}{PMLR},
  \bibinfo{pages}{6028--6039}.
\newblock


\bibitem[Lindner et~al\mbox{.}(2026)]%
        {lindner2026zerottc}
\bibfield{author}{\bibinfo{person}{Dominik Lindner}, \bibinfo{person}{Johann
  Schmidt}, \bibinfo{person}{Tom Siegl}, \bibinfo{person}{Martin Becker}, {and}
  \bibinfo{person}{Sebastian Stober}.} \bibinfo{year}{2026}\natexlab{}.
\newblock \showarticletitle{Zero-Shot Test-Time Canonicalization Using
  Out-of-Distribution Scoring}. In \bibinfo{booktitle}{\emph{Machine Learning
  and Knowledge Discovery in Databases. Research Track (ECML PKDD 2026)}}
  \emph{(\bibinfo{series}{Lecture Notes in Computer Science},
  Vol.~\bibinfo{volume}{16946})}. \bibinfo{publisher}{Springer},
  \bibinfo{pages}{168--185}.
\newblock
\href{https://doi.org/10.1007/978-3-032-37673-2_10}{doi:\nolinkurl{10.1007/978-3-032-37673-2_10}}


\bibitem[Lipton et~al\mbox{.}(2018)]%
        {lipton2018bbse}
\bibfield{author}{\bibinfo{person}{Zachary~C. Lipton},
  \bibinfo{person}{Yu-Xiang Wang}, {and} \bibinfo{person}{Alexander~J. Smola}.}
  \bibinfo{year}{2018}\natexlab{}.
\newblock \showarticletitle{Detecting and Correcting for Label Shift with Black
  Box Predictors}. In \bibinfo{booktitle}{\emph{Proceedings of the 35th
  International Conference on Machine Learning (ICML)}}
  \emph{(\bibinfo{series}{Proceedings of Machine Learning Research},
  Vol.~\bibinfo{volume}{80})}. \bibinfo{publisher}{PMLR},
  \bibinfo{pages}{3122--3130}.
\newblock
\urldef\tempurl%
\url{https://proceedings.mlr.press/v80/lipton18a.html}
\showURL{%
\tempurl}


\bibitem[Maia~Polo et~al\mbox{.}(2023)]%
        {polo2023diagnostics}
\bibfield{author}{\bibinfo{person}{Felipe Maia~Polo}, \bibinfo{person}{Rafael
  Izbicki}, \bibinfo{person}{Evanildo~Gomes Lacerda, Jr.},
  \bibinfo{person}{Juan~Pablo Ibieta-Jimenez}, {and} \bibinfo{person}{Renato
  Vicente}.} \bibinfo{year}{2023}\natexlab{}.
\newblock \showarticletitle{A Unified Framework for Dataset Shift Diagnostics}.
\newblock \bibinfo{journal}{\emph{Information Sciences}}  \bibinfo{volume}{649}
  (\bibinfo{year}{2023}), \bibinfo{pages}{119612}.
\newblock
\href{https://doi.org/10.1016/j.ins.2023.119612}{doi:\nolinkurl{10.1016/j.ins.2023.119612}}


\bibitem[Malekzadeh et~al\mbox{.}(2019)]%
        {motionsense}
\bibfield{author}{\bibinfo{person}{Mohammad Malekzadeh},
  \bibinfo{person}{Richard~G. Clegg}, \bibinfo{person}{Andrea Cavallaro}, {and}
  \bibinfo{person}{Hamed Haddadi}.} \bibinfo{year}{2019}\natexlab{}.
\newblock \showarticletitle{Mobile Sensor Data Anonymization}. In
  \bibinfo{booktitle}{\emph{Proceedings of the International Conference on
  Internet of Things Design and Implementation (IoTDI)}}.
  \bibinfo{publisher}{ACM}, \bibinfo{pages}{49--58}.
\newblock
\href{https://doi.org/10.1145/3302505.3310068}{doi:\nolinkurl{10.1145/3302505.3310068}}


\bibitem[Mathur et~al\mbox{.}(2018)]%
        {mathur2018using}
\bibfield{author}{\bibinfo{person}{Akhil Mathur}, \bibinfo{person}{Tianlin
  Zhang}, \bibinfo{person}{Sourav Bhattacharya}, \bibinfo{person}{Petar
  Veli{\v{c}}kovi{\'c}}, \bibinfo{person}{Leonid Joffe},
  \bibinfo{person}{Nicholas~D. Lane}, \bibinfo{person}{Fahim Kawsar}, {and}
  \bibinfo{person}{Pietro Li{\`o}}.} \bibinfo{year}{2018}\natexlab{}.
\newblock \showarticletitle{Using Deep Data Augmentation Training to Address
  Software and Hardware Heterogeneities in Wearable and Smartphone Sensing
  Devices}. In \bibinfo{booktitle}{\emph{2018 17th ACM/IEEE International
  Conference on Information Processing in Sensor Networks (IPSN)}}.
  \bibinfo{publisher}{IEEE}, \bibinfo{pages}{200--211}.
\newblock
\href{https://doi.org/10.1109/IPSN.2018.00048}{doi:\nolinkurl{10.1109/IPSN.2018.00048}}


\bibitem[Mizell(2003)]%
        {mizell2003using}
\bibfield{author}{\bibinfo{person}{David Mizell}.}
  \bibinfo{year}{2003}\natexlab{}.
\newblock \showarticletitle{Using Gravity to Estimate Accelerometer
  Orientation}. In \bibinfo{booktitle}{\emph{Proceedings of the Seventh IEEE
  International Symposium on Wearable Computers (ISWC)}}.
  \bibinfo{publisher}{IEEE}, \bibinfo{pages}{252--253}.
\newblock
\href{https://doi.org/10.1109/ISWC.2003.1241424}{doi:\nolinkurl{10.1109/ISWC.2003.1241424}}


\bibitem[Mondal et~al\mbox{.}(2023)]%
        {mondal2023equivariant}
\bibfield{author}{\bibinfo{person}{Arnab~Kumar Mondal},
  \bibinfo{person}{Siba~Smarak Panigrahi}, \bibinfo{person}{S{\'e}kou-Oumar
  Kaba}, \bibinfo{person}{Sai Rajeswar}, {and} \bibinfo{person}{Siamak
  Ravanbakhsh}.} \bibinfo{year}{2023}\natexlab{}.
\newblock \showarticletitle{Equivariant Adaptation of Large Pretrained Models}.
  In \bibinfo{booktitle}{\emph{Advances in Neural Information Processing
  Systems (NeurIPS)}}, Vol.~\bibinfo{volume}{36}.
  \bibinfo{pages}{50293--50309}.
\newblock


\bibitem[Nguyen et~al\mbox{.}(2025)]%
        {nguyen2025reliably}
\bibfield{author}{\bibinfo{person}{Viet Nguyen}, \bibinfo{person}{Changjian
  Shui}, \bibinfo{person}{Vijay Giri}, \bibinfo{person}{Siddharth Arya},
  \bibinfo{person}{Amol Verma}, \bibinfo{person}{Fahad Razak}, {and}
  \bibinfo{person}{Rahul~G. Krishnan}.} \bibinfo{year}{2025}\natexlab{}.
\newblock \showarticletitle{Reliably Detecting Model Failures in Deployment
  Without Labels}. In \bibinfo{booktitle}{\emph{Advances in Neural Information
  Processing Systems (NeurIPS)}}, Vol.~\bibinfo{volume}{38}.
  \bibinfo{publisher}{Curran Associates, Inc.}, \bibinfo{pages}{7864--7908}.
\newblock
\href{https://doi.org/10.52202/085713-0269}{doi:\nolinkurl{10.52202/085713-0269}}


\bibitem[Niu et~al\mbox{.}(2022)]%
        {niu2022eata}
\bibfield{author}{\bibinfo{person}{Shuaicheng Niu}, \bibinfo{person}{Jiaxiang
  Wu}, \bibinfo{person}{Yifan Zhang}, \bibinfo{person}{Yaofo Chen},
  \bibinfo{person}{Shijian Zheng}, \bibinfo{person}{Peilin Zhao}, {and}
  \bibinfo{person}{Mingkui Tan}.} \bibinfo{year}{2022}\natexlab{}.
\newblock \showarticletitle{Efficient Test-Time Model Adaptation without
  Forgetting}. In \bibinfo{booktitle}{\emph{Proceedings of the 39th
  International Conference on Machine Learning (ICML)}}
  \emph{(\bibinfo{series}{Proceedings of Machine Learning Research},
  Vol.~\bibinfo{volume}{162})}. \bibinfo{publisher}{PMLR},
  \bibinfo{pages}{16888--16905}.
\newblock


\bibitem[Niu et~al\mbox{.}(2023)]%
        {niu2023sar}
\bibfield{author}{\bibinfo{person}{Shuaicheng Niu}, \bibinfo{person}{Jiaxiang
  Wu}, \bibinfo{person}{Yifan Zhang}, \bibinfo{person}{Zhiquan Wen},
  \bibinfo{person}{Yaofo Chen}, \bibinfo{person}{Peilin Zhao}, {and}
  \bibinfo{person}{Mingkui Tan}.} \bibinfo{year}{2023}\natexlab{}.
\newblock \showarticletitle{Towards Stable Test-Time Adaptation in Dynamic Wild
  World}. In \bibinfo{booktitle}{\emph{International Conference on Learning
  Representations (ICLR)}}.
\newblock
\showeprint[arxiv]{2302.12400}


\bibitem[Ord{\'o}{\~n}ez and Roggen(2016)]%
        {ordonez2016deepconvlstm}
\bibfield{author}{\bibinfo{person}{Francisco~Javier Ord{\'o}{\~n}ez} {and}
  \bibinfo{person}{Daniel Roggen}.} \bibinfo{year}{2016}\natexlab{}.
\newblock \showarticletitle{Deep Convolutional and {LSTM} Recurrent Neural
  Networks for Multimodal Wearable Activity Recognition}.
\newblock \bibinfo{journal}{\emph{Sensors}} \bibinfo{volume}{16},
  \bibinfo{number}{1} (\bibinfo{year}{2016}), \bibinfo{pages}{115}.
\newblock
\href{https://doi.org/10.3390/s16010115}{doi:\nolinkurl{10.3390/s16010115}}


\bibitem[Ord{\'o}{\~n}ez~Morales and Roggen(2016)]%
        {morales2016deep}
\bibfield{author}{\bibinfo{person}{Francisco~Javier Ord{\'o}{\~n}ez~Morales}
  {and} \bibinfo{person}{Daniel Roggen}.} \bibinfo{year}{2016}\natexlab{}.
\newblock \showarticletitle{Deep Convolutional Feature Transfer across Mobile
  Activity Recognition Domains, Sensor Modalities and Locations}. In
  \bibinfo{booktitle}{\emph{Proceedings of the 2016 ACM International Symposium
  on Wearable Computers (ISWC)}}. \bibinfo{publisher}{ACM},
  \bibinfo{pages}{92--99}.
\newblock
\href{https://doi.org/10.1145/2971763.2971764}{doi:\nolinkurl{10.1145/2971763.2971764}}


\bibitem[Park et~al\mbox{.}(2024)]%
        {park2024silentimpact}
\bibfield{author}{\bibinfo{person}{Junyong Park}, \bibinfo{person}{Saelyne
  Yang}, {and} \bibinfo{person}{Sungho Jo}.} \bibinfo{year}{2024}\natexlab{}.
\newblock \showarticletitle{Silent Impact: Tracking Tennis Shots from the
  Passive Arm}. In \bibinfo{booktitle}{\emph{Proceedings of the 37th Annual ACM
  Symposium on User Interface Software and Technology (UIST)}}.
  \bibinfo{publisher}{ACM}, \bibinfo{pages}{1--15}.
\newblock
\href{https://doi.org/10.1145/3654777.3676403}{doi:\nolinkurl{10.1145/3654777.3676403}}


\bibitem[Peng et~al\mbox{.}(2019)]%
        {peng2019moment}
\bibfield{author}{\bibinfo{person}{Xingchao Peng}, \bibinfo{person}{Qinxun
  Bai}, \bibinfo{person}{Xide Xia}, \bibinfo{person}{Zijun Huang},
  \bibinfo{person}{Kate Saenko}, {and} \bibinfo{person}{Bo Wang}.}
  \bibinfo{year}{2019}\natexlab{}.
\newblock \showarticletitle{Moment Matching for Multi-Source Domain
  Adaptation}. In \bibinfo{booktitle}{\emph{Proceedings of the IEEE/CVF
  International Conference on Computer Vision (ICCV)}}.
  \bibinfo{publisher}{IEEE}, \bibinfo{pages}{1406--1415}.
\newblock
\href{https://doi.org/10.1109/ICCV.2019.00149}{doi:\nolinkurl{10.1109/ICCV.2019.00149}}


\bibitem[Qin et~al\mbox{.}(2019)]%
        {qin2019cross}
\bibfield{author}{\bibinfo{person}{Xin Qin}, \bibinfo{person}{Yiqiang Chen},
  \bibinfo{person}{Jindong Wang}, {and} \bibinfo{person}{Chaohui Yu}.}
  \bibinfo{year}{2019}\natexlab{}.
\newblock \showarticletitle{Cross-Dataset Activity Recognition via Adaptive
  Spatial-Temporal Transfer Learning}.
\newblock \bibinfo{journal}{\emph{Proceedings of the ACM on Interactive,
  Mobile, Wearable and Ubiquitous Technologies}} \bibinfo{volume}{3},
  \bibinfo{number}{4} (\bibinfo{year}{2019}), \bibinfo{pages}{1--25}.
\newblock
\href{https://doi.org/10.1145/3369818}{doi:\nolinkurl{10.1145/3369818}}


\bibitem[Rabanser et~al\mbox{.}(2019)]%
        {rabanser2019failing}
\bibfield{author}{\bibinfo{person}{Stephan Rabanser}, \bibinfo{person}{Stephan
  G{\"u}nnemann}, {and} \bibinfo{person}{Zachary~C. Lipton}.}
  \bibinfo{year}{2019}\natexlab{}.
\newblock \showarticletitle{Failing Loudly: An Empirical Study of Methods for
  Detecting Dataset Shift}. In \bibinfo{booktitle}{\emph{Advances in Neural
  Information Processing Systems (NeurIPS)}}, Vol.~\bibinfo{volume}{32}.
  \bibinfo{publisher}{Curran Associates, Inc.}, \bibinfo{pages}{1396--1408}.
\newblock
\urldef\tempurl%
\url{https://proceedings.neurips.cc/paper/2019/hash/846c260d715e5b854ffad5f70a516c88-Abstract.html}
\showURL{%
\tempurl}


\bibitem[Ragab et~al\mbox{.}(2023a)]%
        {ragab2023adatime}
\bibfield{author}{\bibinfo{person}{Mohamed Ragab}, \bibinfo{person}{Emadeldeen
  Eldele}, \bibinfo{person}{Wee~Ling Tan}, \bibinfo{person}{Chuan-Sheng Foo},
  \bibinfo{person}{Zhenghua Chen}, \bibinfo{person}{Min Wu},
  \bibinfo{person}{Chee-Keong Kwoh}, {and} \bibinfo{person}{Xiaoli Li}.}
  \bibinfo{year}{2023}\natexlab{a}.
\newblock \showarticletitle{{AdaTime}: A Benchmarking Suite for Domain
  Adaptation on Time Series Data}.
\newblock \bibinfo{journal}{\emph{ACM Transactions on Knowledge Discovery from
  Data}} \bibinfo{volume}{17}, \bibinfo{number}{8}, Article
  \bibinfo{articleno}{106} (\bibinfo{year}{2023}),
  \bibinfo{numpages}{18}~pages.
\newblock
\href{https://doi.org/10.1145/3587937}{doi:\nolinkurl{10.1145/3587937}}


\bibitem[Ragab et~al\mbox{.}(2023b)]%
        {mapu}
\bibfield{author}{\bibinfo{person}{Mohamed Ragab}, \bibinfo{person}{Emadeldeen
  Eldele}, \bibinfo{person}{Min Wu}, \bibinfo{person}{Chuan-Sheng Foo},
  \bibinfo{person}{Xiaoli Li}, {and} \bibinfo{person}{Zhenghua Chen}.}
  \bibinfo{year}{2023}\natexlab{b}.
\newblock \showarticletitle{Source-Free Domain Adaptation with Temporal
  Imputation for Time Series Data}. In \bibinfo{booktitle}{\emph{Proceedings of
  the 29th ACM SIGKDD Conference on Knowledge Discovery and Data Mining
  (KDD)}}. \bibinfo{pages}{1989--1998}.
\newblock
\href{https://doi.org/10.1145/3580305.3599507}{doi:\nolinkurl{10.1145/3580305.3599507}}


\bibitem[Reiss and Stricker(2012)]%
        {pamap2}
\bibfield{author}{\bibinfo{person}{Attila Reiss} {and} \bibinfo{person}{Didier
  Stricker}.} \bibinfo{year}{2012}\natexlab{}.
\newblock \showarticletitle{Introducing a New Benchmarked Dataset for Activity
  Monitoring}. In \bibinfo{booktitle}{\emph{2012 16th International Symposium
  on Wearable Computers (ISWC)}}. \bibinfo{publisher}{IEEE},
  \bibinfo{pages}{108--109}.
\newblock
\href{https://doi.org/10.1109/ISWC.2012.13}{doi:\nolinkurl{10.1109/ISWC.2012.13}}


\bibitem[Rouast et~al\mbox{.}(2020)]%
        {rouast2020oreba}
\bibfield{author}{\bibinfo{person}{Philipp~V. Rouast}, \bibinfo{person}{Hamid
  Heydarian}, \bibinfo{person}{Marc T.~P. Adam}, {and}
  \bibinfo{person}{Megan~E. Rollo}.} \bibinfo{year}{2020}\natexlab{}.
\newblock \showarticletitle{{OREBA}: A Dataset for Objectively Recognizing
  Eating Behavior and Associated Intake}.
\newblock \bibinfo{journal}{\emph{IEEE Access}}  \bibinfo{volume}{8}
  (\bibinfo{year}{2020}), \bibinfo{pages}{181955--181963}.
\newblock
\href{https://doi.org/10.1109/ACCESS.2020.3026965}{doi:\nolinkurl{10.1109/ACCESS.2020.3026965}}


\bibitem[Saeed et~al\mbox{.}(2019)]%
        {saeed2019selfsup}
\bibfield{author}{\bibinfo{person}{Aaqib Saeed}, \bibinfo{person}{Tanir
  Ozcelebi}, {and} \bibinfo{person}{Johan Lukkien}.}
  \bibinfo{year}{2019}\natexlab{}.
\newblock \showarticletitle{Multi-task Self-Supervised Learning for Human
  Activity Detection}.
\newblock \bibinfo{journal}{\emph{Proceedings of the ACM on Interactive,
  Mobile, Wearable and Ubiquitous Technologies (IMWUT)}} \bibinfo{volume}{3},
  \bibinfo{number}{2}, Article \bibinfo{articleno}{61} (\bibinfo{year}{2019}),
  \bibinfo{numpages}{30}~pages.
\newblock
\href{https://doi.org/10.1145/3328932}{doi:\nolinkurl{10.1145/3328932}}


\bibitem[Schirmer et~al\mbox{.}(2025)]%
        {schirmer2025monitoring}
\bibfield{author}{\bibinfo{person}{Mona Schirmer}, \bibinfo{person}{Metod
  Jazbec}, \bibinfo{person}{Christian Andersson~Naesseth}, {and}
  \bibinfo{person}{Eric Nalisnick}.} \bibinfo{year}{2025}\natexlab{}.
\newblock \showarticletitle{Monitoring Risks in Test-Time Adaptation}. In
  \bibinfo{booktitle}{\emph{Advances in Neural Information Processing Systems
  (NeurIPS)}}, Vol.~\bibinfo{volume}{38}. \bibinfo{publisher}{Curran
  Associates, Inc.}, \bibinfo{pages}{80915--80948}.
\newblock
\href{https://doi.org/10.52202/085713-2705}{doi:\nolinkurl{10.52202/085713-2705}}


\bibitem[Schmidt and Stober(2024)]%
        {schmidt2024its}
\bibfield{author}{\bibinfo{person}{Johann Schmidt} {and}
  \bibinfo{person}{Sebastian Stober}.} \bibinfo{year}{2024}\natexlab{}.
\newblock \showarticletitle{Tilt your Head: Activating the Hidden
  Spatial-Invariance of Classifiers}. In \bibinfo{booktitle}{\emph{Proceedings
  of the 41st International Conference on Machine Learning (ICML)}}
  \emph{(\bibinfo{series}{Proceedings of Machine Learning Research},
  Vol.~\bibinfo{volume}{235})}. \bibinfo{pages}{43705--43722}.
\newblock


\bibitem[Schneider et~al\mbox{.}(2020)]%
        {schneider2020bnadapt}
\bibfield{author}{\bibinfo{person}{Steffen Schneider}, \bibinfo{person}{Evgenia
  Rusak}, \bibinfo{person}{Luisa Eck}, \bibinfo{person}{Oliver Bringmann},
  \bibinfo{person}{Wieland Brendel}, {and} \bibinfo{person}{Matthias Bethge}.}
  \bibinfo{year}{2020}\natexlab{}.
\newblock \showarticletitle{Improving Robustness against Common Corruptions by
  Covariate Shift Adaptation}. In \bibinfo{booktitle}{\emph{Advances in Neural
  Information Processing Systems (NeurIPS)}}, Vol.~\bibinfo{volume}{33}.
  \bibinfo{publisher}{Curran Associates, Inc.}, \bibinfo{pages}{11539--11551}.
\newblock


\bibitem[Shanmugam et~al\mbox{.}(2021)]%
        {shanmugam2021tta}
\bibfield{author}{\bibinfo{person}{Divya Shanmugam}, \bibinfo{person}{Davis
  Blalock}, \bibinfo{person}{Guha Balakrishnan}, {and} \bibinfo{person}{John
  Guttag}.} \bibinfo{year}{2021}\natexlab{}.
\newblock \showarticletitle{Better Aggregation in Test-Time Augmentation}. In
  \bibinfo{booktitle}{\emph{Proceedings of the IEEE/CVF International
  Conference on Computer Vision (ICCV)}}. \bibinfo{publisher}{IEEE},
  \bibinfo{pages}{1194--1203}.
\newblock
\href{https://doi.org/10.1109/ICCV48922.2021.00125}{doi:\nolinkurl{10.1109/ICCV48922.2021.00125}}


\bibitem[Sikder and Nahid(2021)]%
        {kuhar}
\bibfield{author}{\bibinfo{person}{Niloy Sikder} {and}
  \bibinfo{person}{Abdullah-Al Nahid}.} \bibinfo{year}{2021}\natexlab{}.
\newblock \showarticletitle{{KU-HAR}: An Open Dataset for Heterogeneous Human
  Activity Recognition}.
\newblock \bibinfo{journal}{\emph{Pattern Recognition Letters}}
  \bibinfo{volume}{146} (\bibinfo{year}{2021}), \bibinfo{pages}{46--54}.
\newblock
\href{https://doi.org/10.1016/j.patrec.2021.02.024}{doi:\nolinkurl{10.1016/j.patrec.2021.02.024}}


\bibitem[Singhal et~al\mbox{.}(2025)]%
        {singhal2025focal}
\bibfield{author}{\bibinfo{person}{Utkarsh Singhal}, \bibinfo{person}{Ryan
  Feng}, \bibinfo{person}{Stella~X. Yu}, {and} \bibinfo{person}{Atul Prakash}.}
  \bibinfo{year}{2025}\natexlab{}.
\newblock \showarticletitle{Test-Time Canonicalization by Foundation Models for
  Robust Perception}. In \bibinfo{booktitle}{\emph{Proceedings of the 42nd
  International Conference on Machine Learning (ICML)}}
  \emph{(\bibinfo{series}{Proceedings of Machine Learning Research},
  Vol.~\bibinfo{volume}{267})}. \bibinfo{pages}{55788--55809}.
\newblock


\bibitem[Stisen et~al\mbox{.}(2015)]%
        {stisen2015smart}
\bibfield{author}{\bibinfo{person}{Allan Stisen}, \bibinfo{person}{Henrik
  Blunck}, \bibinfo{person}{Sourav Bhattacharya}, \bibinfo{person}{Thor~Siiger
  Prentow}, \bibinfo{person}{Mikkel~Baun Kj{\ae}rgaard}, \bibinfo{person}{Anind
  Dey}, \bibinfo{person}{Tobias Sonne}, {and} \bibinfo{person}{Mads~M{\o}ller
  Jensen}.} \bibinfo{year}{2015}\natexlab{}.
\newblock \showarticletitle{Smart Devices are Different: Assessing and
  Mitigating Mobile Sensing Heterogeneities for Activity Recognition}. In
  \bibinfo{booktitle}{\emph{Proceedings of the 13th ACM Conference on Embedded
  Networked Sensor Systems (SenSys)}}. \bibinfo{publisher}{ACM},
  \bibinfo{pages}{127--140}.
\newblock
\href{https://doi.org/10.1145/2809695.2809718}{doi:\nolinkurl{10.1145/2809695.2809718}}


\bibitem[Straczkiewicz et~al\mbox{.}(2019)]%
        {straczkiewicz2019placement}
\bibfield{author}{\bibinfo{person}{Marcin Straczkiewicz},
  \bibinfo{person}{Nancy~W. Glynn}, {and} \bibinfo{person}{Jaroslaw Harezlak}.}
  \bibinfo{year}{2019}\natexlab{}.
\newblock \showarticletitle{On Placement, Location and Orientation of
  Wrist-Worn Tri-Axial Accelerometers during Free-Living Measurements}.
\newblock \bibinfo{journal}{\emph{Sensors}} \bibinfo{volume}{19},
  \bibinfo{number}{9} (\bibinfo{year}{2019}), \bibinfo{pages}{2095}.
\newblock
\href{https://doi.org/10.3390/s19092095}{doi:\nolinkurl{10.3390/s19092095}}


\bibitem[Sun et~al\mbox{.}(2016)]%
        {coral}
\bibfield{author}{\bibinfo{person}{Baochen Sun}, \bibinfo{person}{Jiashi Feng},
  {and} \bibinfo{person}{Kate Saenko}.} \bibinfo{year}{2016}\natexlab{}.
\newblock \showarticletitle{Return of Frustratingly Easy Domain Adaptation}. In
  \bibinfo{booktitle}{\emph{Proceedings of the AAAI Conference on Artificial
  Intelligence}}, Vol.~\bibinfo{volume}{30}. \bibinfo{pages}{2058--2065}.
\newblock
\href{https://doi.org/10.1609/aaai.v30i1.10306}{doi:\nolinkurl{10.1609/aaai.v30i1.10306}}


\bibitem[Sztyler and Stuckenschmidt(2016)]%
        {sztyler2016onbody}
\bibfield{author}{\bibinfo{person}{Timo Sztyler} {and} \bibinfo{person}{Heiner
  Stuckenschmidt}.} \bibinfo{year}{2016}\natexlab{}.
\newblock \showarticletitle{On-Body Localization of Wearable Devices: An
  Investigation of Position-Aware Activity Recognition}. In
  \bibinfo{booktitle}{\emph{2016 IEEE International Conference on Pervasive
  Computing and Communications (PerCom)}}. \bibinfo{publisher}{IEEE},
  \bibinfo{pages}{1--9}.
\newblock
\href{https://doi.org/10.1109/PERCOM.2016.7456521}{doi:\nolinkurl{10.1109/PERCOM.2016.7456521}}


\bibitem[Tang et~al\mbox{.}(2020)]%
        {tang2020simclrhar}
\bibfield{author}{\bibinfo{person}{Chi~Ian Tang}, \bibinfo{person}{Ignacio
  Perez-Pozuelo}, \bibinfo{person}{Dimitris Spathis}, {and}
  \bibinfo{person}{Cecilia Mascolo}.} \bibinfo{year}{2020}\natexlab{}.
\newblock \bibinfo{title}{Exploring Contrastive Learning in Human Activity
  Recognition for Healthcare}.
\newblock
\showeprint[arxiv]{2011.11542}


\bibitem[Thiemjarus(2010)]%
        {thiemjarus2010device}
\bibfield{author}{\bibinfo{person}{Surapa Thiemjarus}.}
  \bibinfo{year}{2010}\natexlab{}.
\newblock \showarticletitle{A Device-Orientation Independent Method for
  Activity Recognition}. In \bibinfo{booktitle}{\emph{2010 International
  Conference on Body Sensor Networks (BSN)}}. \bibinfo{publisher}{IEEE},
  \bibinfo{pages}{19--23}.
\newblock
\href{https://doi.org/10.1109/BSN.2010.55}{doi:\nolinkurl{10.1109/BSN.2010.55}}


\bibitem[Twaites et~al\mbox{.}(2019)]%
        {twaites2019wrist}
\bibfield{author}{\bibinfo{person}{Joshua Twaites}, \bibinfo{person}{Richard
  Everson}, \bibinfo{person}{Joss Langford}, {and} \bibinfo{person}{Melvyn
  Hillsdon}.} \bibinfo{year}{2019}\natexlab{}.
\newblock \showarticletitle{Achieving Accelerometer Wrist and Orientation
  Invariance in Physical Activity Classification via Domain Adaption}.
\newblock \bibinfo{journal}{\emph{Journal for the Measurement of Physical
  Behaviour}} \bibinfo{volume}{2}, \bibinfo{number}{4} (\bibinfo{year}{2019}),
  \bibinfo{pages}{256--262}.
\newblock
\href{https://doi.org/10.1123/jmpb.2018-0058}{doi:\nolinkurl{10.1123/jmpb.2018-0058}}


\bibitem[Um et~al\mbox{.}(2017)]%
        {um2017augmentation}
\bibfield{author}{\bibinfo{person}{Terry~T. Um}, \bibinfo{person}{Franz M.~J.
  Pfister}, \bibinfo{person}{Daniel Pichler}, \bibinfo{person}{Satoshi Endo},
  \bibinfo{person}{Muriel Lang}, \bibinfo{person}{Sandra Hirche},
  \bibinfo{person}{Urban Fietzek}, {and} \bibinfo{person}{Dana Kuli{\'c}}.}
  \bibinfo{year}{2017}\natexlab{}.
\newblock \showarticletitle{Data Augmentation of Wearable Sensor Data for
  Parkinson's Disease Monitoring Using Convolutional Neural Networks}. In
  \bibinfo{booktitle}{\emph{Proceedings of the 19th ACM International
  Conference on Multimodal Interaction (ICMI)}}. \bibinfo{publisher}{ACM},
  \bibinfo{pages}{216--220}.
\newblock
\href{https://doi.org/10.1145/3136755.3136817}{doi:\nolinkurl{10.1145/3136755.3136817}}


\bibitem[Wang et~al\mbox{.}(2026)]%
        {wang2026eating}
\bibfield{author}{\bibinfo{person}{Chunzhuo Wang}, \bibinfo{person}{Emma
  De~Schuyteneer}, \bibinfo{person}{Huaidong Du}, \bibinfo{person}{Aiden
  Doherty}, \bibinfo{person}{Elske Vrieze}, {and} \bibinfo{person}{Bart
  Vanrumste}.} \bibinfo{year}{2026}\natexlab{}.
\newblock \bibinfo{title}{Generalizable Framework of Eating Episode Detection
  on Free-Living Wrist-Worn Wearable Data}.
\newblock
\showeprint[arxiv]{2607.01376}


\bibitem[Wang et~al\mbox{.}(2021)]%
        {tent}
\bibfield{author}{\bibinfo{person}{Dequan Wang}, \bibinfo{person}{Evan
  Shelhamer}, \bibinfo{person}{Shaoteng Liu}, \bibinfo{person}{Bruno
  Olshausen}, {and} \bibinfo{person}{Trevor Darrell}.}
  \bibinfo{year}{2021}\natexlab{}.
\newblock \showarticletitle{Tent: Fully Test-Time Adaptation by Entropy
  Minimization}. In \bibinfo{booktitle}{\emph{International Conference on
  Learning Representations (ICLR)}}.
\newblock
\showeprint[arxiv]{2006.10726}


\bibitem[Wang et~al\mbox{.}(2019)]%
        {wang2019harsurvey}
\bibfield{author}{\bibinfo{person}{Jindong Wang}, \bibinfo{person}{Yiqiang
  Chen}, \bibinfo{person}{Shuji Hao}, \bibinfo{person}{Xiaohui Peng}, {and}
  \bibinfo{person}{Lisha Hu}.} \bibinfo{year}{2019}\natexlab{}.
\newblock \showarticletitle{Deep Learning for Sensor-Based Activity
  Recognition: A Survey}.
\newblock \bibinfo{journal}{\emph{Pattern Recognition Letters}}
  \bibinfo{volume}{119} (\bibinfo{year}{2019}), \bibinfo{pages}{3--11}.
\newblock
\href{https://doi.org/10.1016/j.patrec.2018.02.010}{doi:\nolinkurl{10.1016/j.patrec.2018.02.010}}


\bibitem[Wang et~al\mbox{.}(2023)]%
        {oftta}
\bibfield{author}{\bibinfo{person}{Shuoyuan Wang}, \bibinfo{person}{Jindong
  Wang}, \bibinfo{person}{Huajun Xi}, \bibinfo{person}{Bob Zhang},
  \bibinfo{person}{Lei Zhang}, {and} \bibinfo{person}{Hongxin Wei}.}
  \bibinfo{year}{2023}\natexlab{}.
\newblock \showarticletitle{Optimization-Free Test-Time Adaptation for
  Cross-Person Activity Recognition}.
\newblock \bibinfo{journal}{\emph{Proceedings of the ACM on Interactive,
  Mobile, Wearable and Ubiquitous Technologies}} \bibinfo{volume}{7},
  \bibinfo{number}{4} (\bibinfo{year}{2023}), \bibinfo{pages}{1--27}.
\newblock
\href{https://doi.org/10.1145/3631450}{doi:\nolinkurl{10.1145/3631450}}


\bibitem[Weiler and Cesa(2019)]%
        {weiler2019e2}
\bibfield{author}{\bibinfo{person}{Maurice Weiler} {and}
  \bibinfo{person}{Gabriele Cesa}.} \bibinfo{year}{2019}\natexlab{}.
\newblock \showarticletitle{General {E(2)}-Equivariant Steerable {CNN}s}. In
  \bibinfo{booktitle}{\emph{Advances in Neural Information Processing Systems
  (NeurIPS)}}, Vol.~\bibinfo{volume}{32}.
\newblock
\urldef\tempurl%
\url{https://proceedings.neurips.cc/paper/2019/hash/45d6637b718d0f24a237069fe41b0db4-Abstract.html}
\showURL{%
\tempurl}


\bibitem[Wilcoxon(1945)]%
        {wilcoxon1945}
\bibfield{author}{\bibinfo{person}{Frank Wilcoxon}.}
  \bibinfo{year}{1945}\natexlab{}.
\newblock \showarticletitle{Individual Comparisons by Ranking Methods}.
\newblock \bibinfo{journal}{\emph{Biometrics Bulletin}} \bibinfo{volume}{1},
  \bibinfo{number}{6} (\bibinfo{year}{1945}), \bibinfo{pages}{80--83}.
\newblock
\href{https://doi.org/10.2307/3001968}{doi:\nolinkurl{10.2307/3001968}}


\bibitem[Wu et~al\mbox{.}(2013)]%
        {wu2013selfcal}
\bibfield{author}{\bibinfo{person}{Xiaoxu Wu}, \bibinfo{person}{Yan Wang},
  \bibinfo{person}{Chieh Chien}, {and} \bibinfo{person}{Greg Pottie}.}
  \bibinfo{year}{2013}\natexlab{}.
\newblock \showarticletitle{Self-Calibration of Sensor Misplacement Based on
  Motion Signatures}. In \bibinfo{booktitle}{\emph{2013 IEEE International
  Conference on Body Sensor Networks (BSN)}}. \bibinfo{publisher}{IEEE},
  \bibinfo{pages}{1--5}.
\newblock
\href{https://doi.org/10.1109/BSN.2013.6575504}{doi:\nolinkurl{10.1109/BSN.2013.6575504}}


\bibitem[Wu and He(2018)]%
        {groupnorm}
\bibfield{author}{\bibinfo{person}{Yuxin Wu} {and} \bibinfo{person}{Kaiming
  He}.} \bibinfo{year}{2018}\natexlab{}.
\newblock \showarticletitle{Group Normalization}. In
  \bibinfo{booktitle}{\emph{Proceedings of the European Conference on Computer
  Vision (ECCV)}}. \bibinfo{pages}{3--19}.
\newblock
\href{https://doi.org/10.1007/978-3-030-01261-8_1}{doi:\nolinkurl{10.1007/978-3-030-01261-8_1}}


\bibitem[Xu et~al\mbox{.}(2023)]%
        {xu2023practically}
\bibfield{author}{\bibinfo{person}{Huatao Xu}, \bibinfo{person}{Pengfei Zhou},
  \bibinfo{person}{Rui Tan}, {and} \bibinfo{person}{Mo Li}.}
  \bibinfo{year}{2023}\natexlab{}.
\newblock \showarticletitle{Practically Adopting Human Activity Recognition}.
  In \bibinfo{booktitle}{\emph{Proceedings of the 29th Annual International
  Conference on Mobile Computing and Networking (MobiCom)}}.
  \bibinfo{publisher}{ACM}, \bibinfo{pages}{1--15}.
\newblock
\href{https://doi.org/10.1145/3570361.3613299}{doi:\nolinkurl{10.1145/3570361.3613299}}


\bibitem[Xu et~al\mbox{.}(2021)]%
        {xu2021limubert}
\bibfield{author}{\bibinfo{person}{Huatao Xu}, \bibinfo{person}{Pengfei Zhou},
  \bibinfo{person}{Rui Tan}, \bibinfo{person}{Mo Li}, {and}
  \bibinfo{person}{Guobin Shen}.} \bibinfo{year}{2021}\natexlab{}.
\newblock \showarticletitle{{LIMU-BERT}: Unleashing the Potential of Unlabeled
  Data for {IMU} Sensing Applications}. In
  \bibinfo{booktitle}{\emph{Proceedings of the 19th ACM Conference on Embedded
  Networked Sensor Systems (SenSys)}}. \bibinfo{publisher}{ACM},
  \bibinfo{pages}{220--233}.
\newblock
\href{https://doi.org/10.1145/3485730.3485937}{doi:\nolinkurl{10.1145/3485730.3485937}}


\bibitem[Yurtman and Barshan(2017)]%
        {yurtman2017orientinv}
\bibfield{author}{\bibinfo{person}{Aras Yurtman} {and} \bibinfo{person}{Billur
  Barshan}.} \bibinfo{year}{2017}\natexlab{}.
\newblock \showarticletitle{Activity Recognition Invariant to Sensor
  Orientation with Wearable Motion Sensors}.
\newblock \bibinfo{journal}{\emph{Sensors}} \bibinfo{volume}{17},
  \bibinfo{number}{8} (\bibinfo{year}{2017}), \bibinfo{pages}{1838}.
\newblock
\href{https://doi.org/10.3390/s17081838}{doi:\nolinkurl{10.3390/s17081838}}


\bibitem[Yurtman et~al\mbox{.}(2018)]%
        {yurtman2018orient}
\bibfield{author}{\bibinfo{person}{Aras Yurtman}, \bibinfo{person}{Billur
  Barshan}, {and} \bibinfo{person}{Bar{\i}\c{s} Fidan}.}
  \bibinfo{year}{2018}\natexlab{}.
\newblock \showarticletitle{Activity Recognition Invariant to Wearable Sensor
  Unit Orientation Using Differential Rotational Transformations Represented by
  Quaternions}.
\newblock \bibinfo{journal}{\emph{Sensors}} \bibinfo{volume}{18},
  \bibinfo{number}{8}, Article \bibinfo{articleno}{2725}
  (\bibinfo{year}{2018}).
\newblock
\href{https://doi.org/10.3390/s18082725}{doi:\nolinkurl{10.3390/s18082725}}


\bibitem[Zhang and Sawchuk(2012)]%
        {uschad}
\bibfield{author}{\bibinfo{person}{Mi Zhang} {and}
  \bibinfo{person}{Alexander~A. Sawchuk}.} \bibinfo{year}{2012}\natexlab{}.
\newblock \showarticletitle{{USC-HAD}: A Daily Activity Dataset for Ubiquitous
  Activity Recognition Using Wearable Sensors}. In
  \bibinfo{booktitle}{\emph{Proceedings of the 2012 ACM Conference on
  Ubiquitous Computing (UbiComp)}}. \bibinfo{publisher}{ACM},
  \bibinfo{pages}{1036--1043}.
\newblock
\href{https://doi.org/10.1145/2370216.2370438}{doi:\nolinkurl{10.1145/2370216.2370438}}


\end{thebibliography}

\clearpage
\appendix

\section{Identification Details}
\label{app:injected}

Identification (Section~\ref{sec:method:identify}) uses five thresholds. The input is left unchanged when the user's
discrepancy is at most the 90th percentile of the discrepancies between each training user and the remaining ones. It is
also left unchanged when the largest held-out reduction is below 0.05 and exceeds the second largest by less than 0.05.
Among the families whose reduction lies within 0.03 of the largest, REFIT names the least flexible one, in the order
rotation and reflection (equal rank, the larger reduction wins), permutation, sign flip, gain and offset. Finally, the
input is left unchanged when the named family leaves a held-out discrepancy above five times the 90th percentile.
The permutation family assigns the six channels to the training channels by their mean, standard deviation, spectral
centroid and autocorrelation, one to one. Gains are clipped to between 0.2 and 5, and the autocorrelations of the
statistics vector use lags 1 and 5.

Table~\ref{tab:injected} lists the shifts injected into the unchanged unit of each held-out subject for the
identification study of Section~\ref{sec:exp:identify}. Rotations and reflections act on the accelerometer and the
gyroscope as in Eq.~\eqref{eq:shift}. The other families act on the raw channels.

\begin{table}[h]
\caption{Injected shifts of the identification study. Gains and offsets alternate in direction over the six channels.}
\label{tab:injected}
\scriptsize
\setlength{\tabcolsep}{0pt}
\begin{tabular*}{\linewidth}{@{\extracolsep{\fill}}lll}
\toprule
Family & Mild & Strong \\
\midrule
Rotation & axis-aligned re-attachment, strap misalignment up to 20 degrees & uniformly random rotation \\
Reflection & side-switch mirror of Eq.~\eqref{eq:mirror} & random member of the reflection family \\
Permutation & accelerometer $x$ and $y$ axes swapped & accelerometer and gyroscope swapped \\
Sign flip & accelerometer $x$ axis & all three gyroscope axes \\
Gain & factors 1.4 and 1/1.4 & factors 2.2 and 1/2.2 \\
Offset & 0.5 training standard deviations & 1.5 training standard deviations \\
\bottomrule
\end{tabular*}
\end{table}

\FloatBarrier
\section{Computational Cost}
\label{app:cost}

Table~\ref{tab:cost} lists the time each method needs on one CPU thread to turn the raw windows of one subject into
predictions, for the three datasets on which Section~\ref{sec:exp:cost} reports the cost.

\begin{table}[!ht]
\caption{Wall-clock time in seconds on one CPU thread for one held-out subject, from raw windows to predictions; median of three runs. Windows is the number of windows; rotation selection times the selection step alone. T3A, LAME and OFTTA process the windows in chunks, the other methods in one batch.}
\label{tab:cost}
\scriptsize
\setlength{\tabcolsep}{0pt}
\begin{tabular*}{\linewidth}{@{\extracolsep{\fill}}lccccccccccc}
\toprule
Dataset & Windows & \shortstack{Inference\\only} & AdaBN & TENT & CORAL & SHOT & T3A & LAME & OFTTA & \shortstack{Rotation\\selection only} & REFIT (ours) \\
\midrule
DSADS & 1,140 & 0.856 & 1.23 & 3.93 & 0.881 & 31.5 & 0.425 & 0.447 & 0.729 & 0.0762 & 1.27 \\
HuGaDB & 1,921 & 0.498 & 0.873 & 2.04 & 0.498 & 25.7 & 0.339 & 0.383 & 0.595 & 0.0719 & 0.961 \\
OPPORTUNITY & 11,110 & 1.55 & 2.63 & 2.44 & 1.32 & 87.6 & 1.22 & 1.40 & 2.09 & 0.154 & 3.02 \\
\bottomrule
\end{tabular*}
\end{table}

\FloatBarrier
\section{Amount of Unlabeled Data}
\label{app:data}

For Table~\ref{tab:data}, REFIT computes the rotation selection and the normalization statistics from a limited
number of the subject's windows, drawn at random or as one continuous stretch, and applies the result to all of them.

\begin{table}[!ht]
\caption{Accuracy under axis-aligned re-attachment when the calibration uses $b$ unlabeled windows of the held-out subject, drawn at random or as one continuous stretch, and is applied to all of them.}
\label{tab:data}
\scriptsize
\setlength{\tabcolsep}{0pt}
\begin{tabular*}{\linewidth}{@{\extracolsep{\fill}}lccccccc}
\toprule
& \multicolumn{3}{c}{Random} & \multicolumn{3}{c}{Continuous} & \\
\cmidrule(lr){2-4}\cmidrule(lr){5-7}
Dataset & $b = 16$ & 64 & 256 & $b = 16$ & 64 & 256 & All \\
\midrule
DSADS & 0.780 & 0.836 & 0.851 & 0.261 & 0.285 & 0.501 & 0.854 \\
HuGaDB & 0.931 & 0.954 & 0.959 & 0.589 & 0.736 & 0.844 & 0.960 \\
OPPORTUNITY & 0.813 & 0.853 & 0.862 & 0.512 & 0.656 & 0.734 & 0.864 \\
\bottomrule
\end{tabular*}
\end{table}

\FloatBarrier
\section{Every Baseline under Simulated Re-attachment}
\label{app:baselines}

Tables~\ref{tab:baselines_aligned} and~\ref{tab:baselines_arbitrary} list the accuracy of every method on each of the eight datasets under
axis-aligned and arbitrary re-attachment, which Figure~\ref{fig:reattach} summarizes. REFIT is the most accurate method on
every dataset under both kinds of re-attachment.

\begin{table}[!ht]
\caption{Accuracy under simulated axis-aligned re-attachment with strap misalignment; leave one subject out. Best in bold.}
\label{tab:baselines_aligned}
\scriptsize
\setlength{\tabcolsep}{0pt}
\begin{tabular*}{\linewidth}{@{\extracolsep{\fill}}lccccccccc}
\toprule
Method & DSADS & HuGaDB & OPPORTUNITY & UCI-HAR & MHEALTH & PAMAP2 & MotionSense & USC-HAD & Mean \\
\midrule
Source model & 0.0870 & 0.214 & 0.274 & 0.248 & 0.106 & 0.121 & 0.399 & 0.206 & 0.207 \\
AdaBN & 0.178 & 0.575 & 0.407 & 0.448 & 0.261 & 0.241 & 0.571 & 0.309 & 0.374 \\
TENT & 0.177 & 0.595 & 0.414 & 0.455 & 0.268 & 0.255 & 0.585 & 0.323 & 0.384 \\
CORAL & 0.120 & 0.311 & 0.429 & 0.308 & 0.181 & 0.149 & 0.445 & 0.237 & 0.272 \\
SHOT & 0.167 & 0.436 & 0.242 & 0.452 & 0.237 & 0.211 & 0.494 & 0.302 & 0.318 \\
T3A & 0.0918 & 0.227 & 0.253 & 0.259 & 0.115 & 0.135 & 0.426 & 0.223 & 0.216 \\
LAME & 0.0835 & 0.208 & 0.282 & 0.247 & 0.107 & 0.117 & 0.399 & 0.205 & 0.206 \\
OFTTA & 0.168 & 0.548 & 0.370 & 0.444 & 0.245 & 0.236 & 0.563 & 0.312 & 0.361 \\
\midrule
REFIT (ours) & \textbf{0.854} & \textbf{0.960} & \textbf{0.864} & \textbf{0.943} & \textbf{0.851} & \textbf{0.720} & \textbf{0.948} & \textbf{0.746} & \textbf{0.861} \\
\bottomrule
\end{tabular*}
\end{table}

\begin{table}[!ht]
\caption{Accuracy under simulated arbitrary re-attachment; leave one subject out. Best in bold.}
\label{tab:baselines_arbitrary}
\scriptsize
\setlength{\tabcolsep}{0pt}
\begin{tabular*}{\linewidth}{@{\extracolsep{\fill}}lccccccccc}
\toprule
Method & DSADS & HuGaDB & OPPORTUNITY & UCI-HAR & MHEALTH & PAMAP2 & MotionSense & USC-HAD & Mean \\
\midrule
Source model & 0.0867 & 0.205 & 0.252 & 0.245 & 0.106 & 0.103 & 0.367 & 0.189 & 0.194 \\
AdaBN & 0.167 & 0.567 & 0.427 & 0.466 & 0.249 & 0.236 & 0.550 & 0.299 & 0.370 \\
TENT & 0.166 & 0.587 & 0.432 & 0.474 & 0.248 & 0.249 & 0.566 & 0.311 & 0.379 \\
CORAL & 0.112 & 0.314 & 0.426 & 0.312 & 0.182 & 0.137 & 0.428 & 0.221 & 0.266 \\
SHOT & 0.146 & 0.442 & 0.222 & 0.468 & 0.222 & 0.193 & 0.456 & 0.292 & 0.305 \\
T3A & 0.0949 & 0.222 & 0.230 & 0.262 & 0.119 & 0.113 & 0.390 & 0.212 & 0.205 \\
LAME & 0.0823 & 0.196 & 0.231 & 0.245 & 0.105 & 0.101 & 0.368 & 0.186 & 0.189 \\
OFTTA & 0.159 & 0.539 & 0.390 & 0.465 & 0.241 & 0.230 & 0.536 & 0.302 & 0.358 \\
\midrule
REFIT (ours) & \textbf{0.468} & \textbf{0.792} & \textbf{0.687} & \textbf{0.801} & \textbf{0.539} & \textbf{0.451} & \textbf{0.802} & \textbf{0.577} & \textbf{0.640} \\
\bottomrule
\end{tabular*}
\end{table}

\FloatBarrier
\section{Identification per Dataset}
\label{app:perds}

Table~\ref{tab:identify_ds} reports identification on each of the nine datasets. The share of cases named correctly
ranges from 0.656 to 0.847, and REFIT is the most accurate of the compared methods on every dataset.

\begin{table}[!ht]
\caption{Identification per dataset: share of cases named correctly and accuracy after the correction. Test cases is the number of cases in the accuracy columns; the share named correctly leaves out the real placement shifts. Best in bold.}
\label{tab:identify_ds}
\scriptsize
\setlength{\tabcolsep}{0pt}
\begin{tabular*}{\linewidth}{@{\extracolsep{\fill}}lcccccc}
\toprule
& & & \multicolumn{4}{c}{Accuracy} \\
\cmidrule(lr){4-7}
Dataset & Test cases & Named correctly & AdaBN & TENT & Always rotation & REFIT (ours) \\
\midrule
DSADS & 1,656 & 0.835 & 0.456 & 0.459 & 0.486 & \textbf{0.717} \\
HuGaDB & 4,536 & 0.777 & 0.792 & 0.807 & 0.804 & \textbf{0.929} \\
MHEALTH & 780 & 0.847 & 0.444 & 0.452 & 0.411 & \textbf{0.652} \\
MotionSense & 936 & 0.729 & 0.719 & 0.735 & 0.786 & \textbf{0.896} \\
OPPORTUNITY & 672 & 0.718 & 0.622 & 0.634 & 0.644 & \textbf{0.756} \\
PAMAP2 & 936 & 0.735 & 0.513 & 0.525 & 0.502 & \textbf{0.695} \\
REALDISP & 6,861 & 0.722 & 0.510 & 0.518 & 0.488 & \textbf{0.736} \\
UCI-HAR & 1,170 & 0.691 & 0.684 & 0.691 & 0.818 & \textbf{0.891} \\
USC-HAD & 546 & 0.656 & 0.516 & 0.529 & 0.502 & \textbf{0.661} \\
\bottomrule
\end{tabular*}
\end{table}

\FloatBarrier
\section{Accelerometers Without Gravity}
\label{app:gravity}

Some phones and datasets deliver linear acceleration, from which gravity has already been subtracted. The mean
acceleration in Eq.~\eqref{eq:discrep} is then a small residual that no longer points along gravity, and matching it
would turn units that were not moved. We detect such units by their gravity share
$r_u = \lVert \mu_u \rVert^2 / (\lVert \mu_u \rVert^2 + \mathrm{tr}\,\Sigma^{a}_u)$, the part of the accelerometer's
second moment that its mean carries. This share does not change under rotation. It is at least 0.207 for
every unit of the raw accelerometers in Section~\ref{sec:exp:reattach} and at most 0.0246 for the
gravity-free streams below. A unit counts as gravity-free when the share of the training windows or of the user's
windows is below 0.05. For such a unit the selection drops the mean term of Eq.~\eqref{eq:discrep}. It keeps the
identity unless the two covariance terms, evaluated at the identity, exceed their 90th percentile over the training
users, where each training user is compared with the remaining ones. Otherwise it selects the rotation that minimizes
the two covariance terms. Normalization re-estimation follows as before. The percentile was chosen on DSADS, HuGaDB,
OPPORTUNITY and REALDISP after removing the mean acceleration of every window. No unit of
Section~\ref{sec:exp:reattach} is gravity-free, so the selection there is exactly Eq.~\eqref{eq:select}.

We evaluate the rule on four gravity-free streams. MotionSense~\cite{motionsense} provides, besides the total acceleration used
in Section~\ref{sec:exp:reattach}, the linear acceleration computed by the phone. UCI-HAR~\cite{ucihar} likewise provides the body acceleration that its authors separate from gravity with a
0.3~Hz filter. From USC-HAD~\cite{uschad} we remove the component below 0.3~Hz in the same way. These give 68 held-out
subjects under the protocol of Section~\ref{sec:exp:reattach}. KU-HAR~\cite{kuhar} records a waist-worn phone whose
accelerometer stream has gravity removed. We use its 57 subjects who perform at least half of the
18 activities and hold them out in ten folds. Every unit of these streams is gravity-free, and
Table~\ref{tab:gravity} reports the accuracy of every method.

\begin{table}[!ht]
\caption{Accuracy for held-out subjects whose accelerometer stream has gravity removed; mean over subjects. Best in bold.}
\label{tab:gravity}
\scriptsize
\setlength{\tabcolsep}{0pt}
\begin{tabular*}{\linewidth}{@{\extracolsep{\fill}}lcccccc}
\toprule
 & \multicolumn{3}{c}{MotionSense, USC-HAD, UCI-HAR (68 subjects)} & \multicolumn{3}{c}{KU-HAR (57 subjects)} \\
\cmidrule(lr){2-4} \cmidrule(lr){5-7}
Method & Axis-aligned & Arbitrary & No re-attachment & Axis-aligned & Arbitrary & No re-attachment \\
\midrule
Source model & 0.400 & 0.396 & 0.787 & 0.379 & 0.352 & 0.792 \\
AdaBN & 0.459 & 0.462 & \textbf{0.880} & 0.405 & 0.386 & \textbf{0.925} \\
TENT & 0.459 & 0.459 & 0.832 & 0.376 & 0.352 & 0.801 \\
CORAL & 0.425 & 0.427 & 0.804 & 0.366 & 0.342 & 0.765 \\
SHOT & 0.459 & 0.460 & 0.855 & 0.383 & 0.365 & 0.865 \\
T3A & 0.421 & 0.419 & 0.812 & 0.401 & 0.375 & 0.823 \\
LAME & 0.395 & 0.391 & 0.768 & 0.366 & 0.335 & 0.769 \\
OFTTA & 0.433 & 0.435 & 0.818 & 0.374 & 0.348 & 0.826 \\
\midrule
REFIT (ours) & \textbf{0.685} & \textbf{0.555} & 0.866 & \textbf{0.783} & \textbf{0.569} & 0.915 \\
\bottomrule
\end{tabular*}
\end{table}

With this rule REFIT is again the best method under both kinds of re-attachment. Under axis-aligned re-attachment it is
more accurate than AdaBN for 62 of the 68 subjects and for all 57 KU-HAR subjects. When no unit was re-attached, the
rule keeps the identity for all but three of the 68 subjects and two of the KU-HAR subjects. REFIT is then second only
to AdaBN and on average 0.0140 below it (0.0102 on KU-HAR). The difference comes entirely from the accuracy of these few turned subjects.

\FloatBarrier
\section{Limitations and Future Work}
\label{app:limits}

\head{Units turned without a re-attachment.}
The selection compares a user's moments with those of the training population and cannot tell a turned unit from a user
whose movements differ strongly from that population. For 18 subjects of the simulated benchmark it turns a unit that
was not moved, and for 9 of them this lowers the accuracy by between 0.0152 and 0.494.
On the gravity-free streams of Appendix~\ref{app:gravity}, the few subjects it turns lose accuracy as well.
Deciding more reliably whether a unit has moved is a natural next step.

\head{Arbitrary re-attachment.}
When a unit can be re-attached in any orientation, the 24 candidate rotations only approximate the true one. The
selection then finds the closest candidate for 51.0\% of the units instead of
92.4\% after an axis-aligned re-attachment. Even the closest candidate leaves a residual rotation
of 42.2 degrees in the median and up to 61.7 degrees, which
normalization re-estimation removes only in part (Table~\ref{tab:baselines_arbitrary}).
Continuous rotation estimates would extend REFIT to arbitrary re-attachment.

\head{Shifts to other body locations.}
No transform of the axes undoes a shift to another body location, because the motion itself differs there. The
correction recovers only 1.41\% of these shifts, and none of the 27 placement shifts in which the REALDISP experimenters
deliberately displaced several units. For such shifts REFIT relies on verification to abstain. Recognizing the new
location and switching to a model for that location~\cite{kunze2005where} would go further.

\head{Simulated re-attachment.}
The simulated benchmark rotates recorded signals and therefore omits the changes of position and skin contact that
come with a real re-attachment. REALDISP is the only public dataset we know of that records the same sensors in an expert placement and after
the subjects re-attach them. Its mutual-displacement
condition, in which the experimenters deliberately displace the sensors, includes only three~subjects.

\head{Recording conditions and devices.}
All datasets record scripted activities in sessions of a few minutes to hours. Moment matching assumes that the user's
windows cover a mix of activities similar to that of the training users. Windows from one continuous stretch, which
cover few activities, need many more of them (Table~\ref{tab:data}). How the moments behave on free-living data,
which is dominated by sitting and lying, remains to be measured. The computational cost was
measured on one thread of a server processor. Absolute times on a watch processor will be longer.

\head{Identification and verification.}
Sign flips are the family that identification names least often (Table~\ref{tab:naming}): REFIT names a
reflection or a rotation for 32.6\% of them and leaves 17.9\% unchanged. Identification is also less reliable on some
datasets, such as USC-HAD (Appendix~\ref{app:perds}). The verification was evaluated for the correction over the 48
signed permutations. After a re-attachment, this correction is less accurate than the correction over the 24 rotations
(Table~\ref{tab:ablation}) and than the correction that identification chooses (Sections~\ref{sec:exp:identify}
and~\ref{sec:exp:verify}). How well the estimate verifies these corrections remains to be measured on real placement
shifts. For the verification study, we had
specified a detector in advance: the held-out residual of Eq.~\eqref{eq:discrep} after the best of the 48 signed
permutations. It decides before the correction and separates the recovered placement shifts with an AUROC of 0.791,
above each estimate before the correction but below each estimate after it (Table~\ref{tab:verify}).

\end{document}